\documentclass[11pt]{article}
\usepackage{graphicx} 
\usepackage[margin=1in]{geometry}
\usepackage[colorlinks, linkcolor=blue, anchorcolor=blue, citecolor=orange]{hyperref}
\usepackage{smile}
\usepackage{xcolor}
\usepackage[sort]{cite}

\newcommand{\diff}{{\rm d}}

\newcommand{\ttXytb}{\tilde{X}_{t}^{y, \leftarrow}}

\newcommand{\bXb}{X^{\leftarrow}}
\newcommand{\bXy}{X^{y, \leftarrow}}

\newcommand{\hatbXb}{\tilde{X}^{\leftarrow}}

\newcommand{\re}[1]{{\color{black} {#1}}}

\title{\bf An Overview of Diffusion Models: Applications, Guided Generation, Statistical Rates and Optimization\thanks{Emails: \texttt{\{minshuochen, jqfan, mengdiw\}@princeton.edu, songmei@berkeley.edu}}}
\author{Minshuo Chen$^1$ \quad Song Mei$^2$ \quad Jianqing Fan$^1$ \quad Mengdi Wang$^1$ \\
$^1$Princeton University \quad $^2$UC Berkeley}
\date{\today}

\begin{document}

\maketitle

\begin{abstract}
Diffusion models, a powerful and universal generative AI technology, have achieved tremendous success in computer vision, audio, reinforcement learning, and computational biology. In these applications, diffusion models provide flexible high-dimensional data modeling, and act as a sampler for generating new samples under active guidance towards task-desired properties. Despite the significant empirical success, theory of diffusion models is very limited, potentially slowing down principled methodological innovations for further harnessing and improving diffusion models. In this paper, we review emerging applications of diffusion models, understanding their sample generation under various controls. Next, we overview the existing theories of diffusion models, covering their statistical properties and sampling capabilities. We adopt a progressive routine, beginning with unconditional diffusion models and connecting to conditional counterparts. Further, we review a new avenue in high-dimensional structured optimization through conditional diffusion models, where searching for solutions is reformulated as a conditional sampling problem and solved by diffusion models. Lastly, we discuss future directions about diffusion models. The purpose of this paper is to provide a well-rounded theoretical exposure for stimulating forward-looking theories and methods of diffusion models.
\end{abstract}

\tableofcontents

\section{Introduction}\label{sec:intro}
The field of artificial intelligence (AI) has been revolutionized by generative models, particularly large language models and diffusion models. Recognized as foundation models \cite{bommasani2021opportunities}, they are trained on massive corpora of data and have opened up vibrant possibilities in machine learning research and applications. While large language models focus on generating coherent text based on context, diffusion models excel at modeling complex data distributions and generating diverse samples, both of which find widespread use across various domains.

Diffusion models, inspired by thermodynamics modeling \cite{sohl2015deep}, have emerged in recent years with ground-breaking performance, surpassing the previous state-of-the-art, such as Generative Adversarial Networks (GANs) \cite{goodfellow2020generative, creswell2018generative} and Variational AutoEncoders (VAEs) \cite{kingma2013auto, kingma2019introduction}. Diffusion models are widely adopted in computer vision and audio generation tasks \cite{song2019generative, dathathri2019plug, ho2020denoising, song2020score, kong2020diffwave, chen2020wavegrad, mittal2021symbolic, huang2022prodiff, jeong2021diff, ulhaq2022efficient, avrahami2022blended, kim2022diffusionclip, bansal2023universal, saharia2022photorealistic, po2023state, zhang2023survey}, and further utilized in text generation \cite{li2022diffusion, yu2022latent, lovelace2022latent}, sequential data modeling \cite{alcaraz2022diffusion, tashiro2021csdi, tevet2022human, tian2023fast}, reinforcement learning and control \cite{pearce2023imitating, chi2023diffusion, hansen2023idql, reuss2023goal, zhu2023diffusion, ding2023consistency}, as well as life-science \cite{cao2022high, chung2022mr, chung2022score, gungor2023adaptive, jing2022torsional, anand2022protein, lee2022proteinsgm, luo2022antigen, mei2022metal, waibel2022diffusion, ingraham2022illuminating, huang20223dlinker, schneuing2022structure, wu2022diffusion, gruver2023protein, weiss2023guided, xu2022geodiff, song2021solving, watson2023novo}. For a more comprehensive exposition of applications, we refer readers to survey papers \cite{yang2022diffusion, li2023diffusion, zhang2023survey, guo2023diffusion, cao2022survey, croitoru2023diffusion}.

The celebrated performance of diffusion models is indispensable to numerous methodological innovations that significantly expand the scope and boost the functionality of diffusion models, enabling high-fidelity generation, efficient sampling, and flexible control of the sample generation. For example, \cite{hoogeboom2021argmax, austin2021structured, gu2022vector, ouyang2023missdiff} extend diffusion models to discrete data generation, while the vanilla diffusion models target at continuous data. Meanwhile, there is an active line of research aiming to expedite the sample generation speed of diffusion models \cite{song2020improved, song2020denoising, rombach2022high, song2023consistency, lu2022dpm, salimans2022progressive, karras2022elucidating, zhang2022gddim, bao2022analytic, liu2023instaflow, zhang2022fast}. Last but not the least, a recent surge of research focuses on fine-tuning diffusion models towards generating samples of desired properties, such as generating images with peculiar aesthetic qualities  \cite{clark2023directly, lee2023aligning, wu2023better, black2023training, fan2023dpok, xu2023imagereward, hao2022optimizing, watson2021learning, wallace2023end}. These task-specific properties are often encoded as guidance to the diffusion model, consisting of conditioning and control signals to steer the sample generation. Notably, guidance allows for the creation of diverse and relevant content across a wide range of applications, which underscores the versatility and adaptability of diffusion models. We term diffusion models with guidance as conditional diffusion models.

Despite the rapidly growing body of empirical advancements, theories of diffusion models fall far behind. Some recent theories view diffusion models as an unsupervised distribution learner and sampler, and thus establish their sampling convergence guarantees \cite{block2020generative, lee2022convergencea, chen2022sampling, lee2022convergenceb, chen2023probability, benton2023linear} and statistical distribution learning guarantees \cite{oko2023diffusion, chen2023score, mei2023deep}. Such results offer invaluable theoretical insights into the efficiency and accuracy of diffusion models for modeling complex data, with a central focus on the unconditioned diffusion models in distribution estimation. This leaves a gap between theory and practice for conditional diffusion models. In specific, a theoretical foundation to support and motivate principled methodologies for guidance design and adapting diffusion models to task-specific needs is still lacking.

This paper serves as a contemporary exposure to diffusion models for stimulating sophisticated and forward-looking study on them. \re{We mainly focus on the following fundamental theoretical questions of diffusion models:
\begin{itemize}
\item Can diffusion models learn data distributions accurately and efficiently? If so, what is the sample complexity, especially for structured data?
\item Can conditional diffusion models generate distributions aligned to guidance? If so, how can we properly design the guidance and what is the sample complexity?
\end{itemize}}
For a systematic study, we will first review how diffusion models work and their emerging applications. Then we provide an overview of the existing theoretical underpinnings pertinent to the questions above. Our ultimate goal is to demonstrate and harness the power of diffusion models, connecting to broad interdisciplinary areas of applied mathematics, statistics, computational biology, and operations research.

\paragraph{Paper Organization} The rest of the paper is organized as follows. In Section~\ref{sec:pre}, we present a continuous-time description of diffusion models using stochastic differential equations. The advantage of the continuous-time point of view lies in the clean and systematic formulation, and the seamless application of discretization schemes to replicate practical implementations.

In Section~\ref{sec:application}, we review emerging applications of diffusion models, especially in various controlled generation tasks, aiming to elucidate the conditional distributions that diffusion models attempt to capture. Then, in Section~\ref{sec:application_opt}, we relate conditional generation to black-box optimization via gauging the quality of the generated samples under control by a reward function.

In Section~\ref{sec:theory_uncondition}, we delve into theoretical preliminaries and review theories of diffusion models. Specifically, in Section~\ref{sec:score_estimation}, we discuss how to learn the score function. Section~\ref{sec:score_guarantee} provides approximation theories for understanding proper neural network architectures for learning the score and statistical sample complexities of estimating the score function. Section~\ref{sec:distro_guarantee} then discusses statistical sample complexities of distribution estimation using diffusion models and sampling theories by viewing diffusion models as samplers.

In Section~\ref{sec:theory_cdm}, we focus on conditional diffusion models, continuing a similar study in Section~\ref{sec:theory_uncondition}. We introduce learning methods of conditional score functions in Section~\ref{sec:conditional_score}, connecting them to unconditional score by a so-called ``guidance'' term. This also motivates fine-tuning methods for conditional diffusion models. Section~\ref{sec:cdm_guarantee} then summarizes unconditional score approximation, estimation, and distribution learning theories. Section~\ref{sec:strength_impact} revisits the guidance in conditional score functions and establishes theoretical insights for the impact of guidance.

In Section~\ref{sec:cdm_opt}, we review theories and methodologies of data-driven black-box optimization using conditional diffusion models. We highlight that diffusion models generate high-fidelity solutions to the optimization objective function, preserving data latent structures, and the quality of the solutions aligns with the optimal off-policy bandit. This opens up new possibilities for optimization in high-dimensional complex and structured
spaces through diffusion models.

Lastly, in Section~\ref{sec:open_question}, we discuss future directions and connections of diffusion models to broad research areas.

\section{Diffusion Model Preliminaries}\label{sec:pre}
\begin{figure}[!htb]
\centering
\includegraphics[width = 0.85\textwidth]{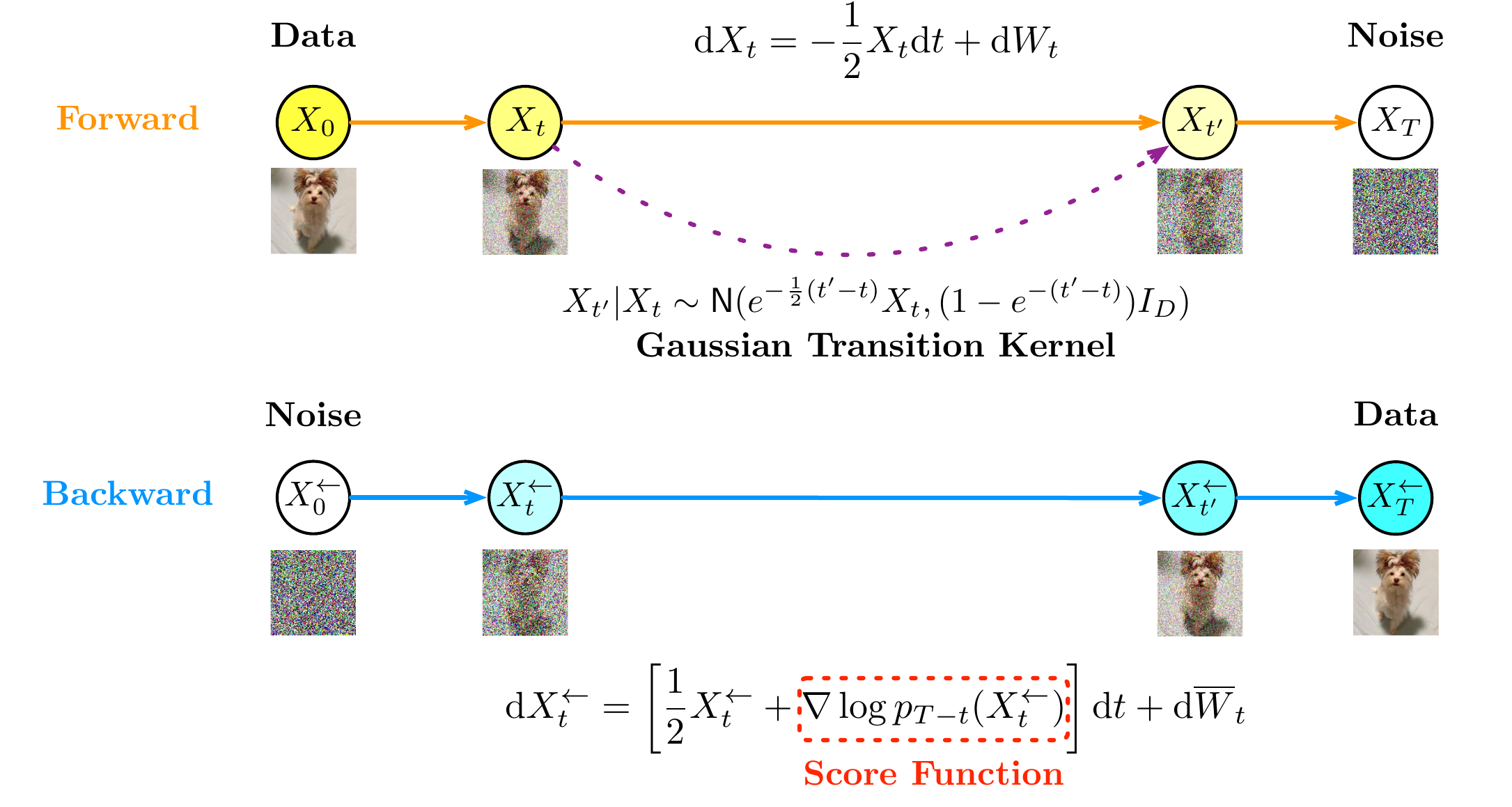}
\caption{Demonstration of forward and backward processes in diffusion models. The forward process is a noise corruption process, while the backward process is used for new sample generation.}
\label{fig:DM}
\end{figure}

Roughly speaking, diffusion model consists of a forward process and a backward process. In the forward process, a clean sample from the data distribution is sequentially corrupted by Gaussian random noise, and in the infinite-time limit, the data distribution is transformed into pure noise. In the backward process, a denoising neural network is trained to sequentially remove the added noise distribution in data and restore new clean data distribution. The forward and backward processes are depicted in Figure~\ref{fig:DM}.

To fully decipher how diffusion models work, we describe the forward and backward processes in a continuous time limit and review how to implement the backward process. Next, we will introduce guidance to realize conditioning in controlled sample generation using conditional diffusion models.

\subsection{Forward and Backward Processes}
The forward process in diffusion models progressively adds noise to the original data. Here we consider the Ornstein-Ulhenbeck process, which is described by the following Stochastic Differential Equation (SDE),
\begin{align}\label{eq:forward_sde}
\diff X_t = -\frac{1}{2} g(t) X_t \diff t + \sqrt{g(t)} \diff W_t ~~~ \text{for} ~~ g(t) > 0,
\end{align}
where initial $X_0 \sim P_{\rm data}$ follows the data distribution, $(W_t)_{t\geq 0}$ is a standard Wiener process, and $g(t)$ is a nondecreasing weighting function. We denote the marginal distribution of $X_t$ at time $t$ as $P_t$. After an infinitesimal time, the forward process \eqref{eq:forward_sde} shrinks the magnitude of data and corrupts data by Gaussian noise.
More precisely, given $X_0$, the conditional distribution of $X_t | X_0$ is Gaussian ${\sf N}(\alpha(t) X_0, h(t)I_D)$, where $\alpha(t) = \exp(-\int_0^t \frac{1}{2}g(s) \diff s)$ and $h(t) = 1 - \alpha^2(t)$. Consequently, under mild conditions, \eqref{eq:forward_sde} transforms initial distribution $P_{\rm data}$ to $P_{\infty} = {\sf N}(0, I_D)$. Therefore, \eqref{eq:forward_sde} is known as the variance preserving forward SDE \cite{song2020score, tang2024score}. 

The value of $g(t)$ controls the noise corruption speed in the forward process. In real-world usage, various choices on $g(t)$ are implemented. \re{One example is to choose $g(t)$ so that the variance of the Gaussian noise in the forward process increases linearly with respect to time \cite{ho2020denoising}}. Later, several improvement techniques of $g(t)$ are proposed, such as a cosine-based variance schedule \cite{nichol2021improved}. To simplify our presentation, we take $g(t) = 1$ for all $t$ in the sequel. 

The forward process \eqref{eq:forward_sde} will terminate at a sufficiently large time $T > 0$, where the corrupted marginal distribution $P_T$ is expected to be close to the standard Gaussian distribution. Then diffusion models generate fake data by reversing the time of \eqref{eq:forward_sde}, which leads to the following backward SDE,
\begin{align}\label{eq:backward_sde}
\diff \bXb_t & = \left[\frac{1}{2}\bXb_t + \nabla \log p_{T-t}(\bXb_t)\right] \diff t + \diff \overline{W}_t \quad \text{for} \quad t \in [0, T),
\end{align}
where $\nabla \log p_t(\cdot)$ is the so-called ``score function'', i.e., the gradient of the log probability density function of $P_t$, \re{$\overline{W}_t$ is another Wiener process independent of $W_t$}, and we use the superscript $\leftarrow$ for distinguishing with the forward process \eqref{eq:forward_sde}. Under mild conditions, when initialized at $\bXb_0 \sim P_T$, the backward process $(\bXb_t)_{0\leq t < T}$ has the same distribution as the forward process $(X_{T-t})_{0 \leq t < T}$ \cite{anderson1982reverse, haussmann1986time}.

Working with \eqref{eq:backward_sde}, however, leads to difficulties, as both the score function $\nabla \log p_t$ and the distribution $P_T$ are unknown. Therefore, several surrogates are deployed in practice. Firstly, we replace the unknown distribution $P_T$ by the standard Gaussian distribution ${\sf N}(0, I_D)$. Secondly, \re{we denote $\hat{s}(x, t)$ as an estimator to the ground truth score function $\nabla \log p_t(x)$. The estimated score $\hat{s}$ is often parameterized by a deep neural network and takes data and time as inputs.} Substituting $\hat{s}$ into the backward process, we obtain the following practical continuous-time backward SDE,
\begin{align}\label{eq:backward_practice}
\diff  {\hatbXb}_t & = \left[\frac{1}{2} {\hatbXb}_t + \hat{s}(\hatbXb_{t}, T-t)\right] \diff t + \diff \overline{W}_t \quad \text{with} \quad \hatbXb_{0} \sim {\sf N}(0, I_D).
\end{align}
Diffusion models then generate data by simulating a discretization of \eqref{eq:backward_practice} with a proper step size. A common practice is to set the step size of order $\cO(1/1000)$ so that the backward SDE \eqref{eq:backward_practice} is discretized to hundreds of steps \cite{ho2020denoising, song2020improved, nichol2021improved}.

It is worth mentioning that simulating the backward process for thousands of steps to generate a sample is time-consuming. Accelerating the sampling speed of diffusion models is an active research direction \cite{song2020improved, song2020denoising, rombach2022high, song2023consistency, lu2022dpm, salimans2022progressive, karras2022elucidating, zhang2022gddim, bao2022analytic, liu2023instaflow, zhang2022fast}. Some notable methods include sampling with stride to reduce the backward steps \cite{nichol2021improved, song2020improved, lu2022dpm}, replacing the backward SDE \eqref{eq:backward_practice} with an ODE or DDIM (Denoising Diffusion Implicit Models) \cite{song2020denoising, karras2022elucidating, zhang2022gddim}, using a pre-trained VAE to extract low-dimensional data representations and then implementing diffusion processes -- known as latent diffusion \cite{rombach2022high}, training distillation and consistency models \cite{song2023consistency, salimans2022progressive, luo2023lcm}, as well as rectified flows \cite{liu2023instaflow}. These methods have found extensive adoption in highly fine-tuned diffusion models, such as Sora and Stable Diffusion \cite{liu2024sora,esser2024scaling}.

\subsection{Conditional Diffusion Models}
Conditional diffusion models generate samples analogous to the unconditioned one, while the major difference is the added conditional information. We denote the conditional information as $y$. Then the goal of conditional diffusion models is to generate samples from the conditional data distribution $P(\cdot | y)$. The conditional forward process is again an Ornstein-Ulhenbeck process:
\begin{align}\label{eq:forward_conditioned}
\diff X_t^y = -\frac{1}{2} X_t^y \diff t + \diff W_t \quad \text{with} \quad X_0^y \sim P_0(\cdot | y) \quad \text{and} \quad t \in (0, T].
\end{align}
Note that the initial distribution is now a conditional distribution $P_0(\cdot | y)$, which is different from the unconditioned forward process \eqref{eq:forward_sde}. The noise corruption is only performed on $x$, while $y$ is kept fixed. We use the superscript $y$ to emphasize the dependence of the process on $y$. Similarly, for sample generation, the backward process reverses the time in \eqref{eq:forward_conditioned}:
\begin{align}
\label{eq:conditional_backward}
\diff \bXy_t & = \left[\frac{1}{2}\bXy_t + \nabla \log p_{T-t}(\bXy_t | y)\right] \diff t + \diff \overline{W}_t \quad \text{for} \quad t \in [0, T).
\end{align}
Here $\nabla \log p_{T-t}(\bXy_t | y)$ is the so-called ``conditional score function'', which replaces the score function in \eqref{eq:backward_sde}. The initialization is identical to \eqref{eq:backward_sde} as $\bXy_0 \sim {\sf N}(0, I_D)$, independent of the guidance $y$. \re{Despite the similarity in forward and backward processes, the major difference between conditional diffusion models and unconditioned ones lies in the estimation of the conditional score function $\nabla \log p_t(\cdot | y)$. In specific, the conditional score function can be related to the unconditioned one, which motivates a collection of practical learning and fine-tuning methods, e.g., classifier guidance and classifier-free guidance \cite{ho2020denoising, song2020score, ho2022classifier}. We defer an in-depth discussion to Section~\ref{sec:conditional_score}}.

With an estimated conditional score function $\hat{s}(x, y, t)$ replacing the ground truth conditional score $\nabla \log p_t(x | y)$, the conditional sample generation is to simulate the following backward process
\begin{align*}
\diff \ttXytb = \left[\frac{1}{2} \tilde{X}_{t}^{y, \leftarrow} + \hat{s}(\tilde{X}_{t}^{y, \leftarrow}, y, T - t) \right] \diff t + \diff \overline{W}_t \quad \text{with} \quad \tilde{X}_0^{y, \leftarrow} \sim {\sf N}(0, I_D).
\end{align*}
In practical implementations, a proper discretization scheme is applied.

\section{Emerging Applications of Diffusion Models}\label{sec:application}
Through extensive developments \cite{song2019generative, ho2020denoising, song2020score, nichol2021improved}, modern diffusion models have achieved a startling success and are implanted in various applications (see, for example, the survey \cite{yang2022diffusion}). We highlight vast applications of diffusion models in the following, with a particular emphasis on conditional diffusion models for controlled sample generation.
 
\subsection{Vision and Audio Generation}
Diffusion models achieve state-of-the-art performance in image and audio generation tasks \cite{song2019generative, dathathri2019plug, ho2020denoising, song2020score, kong2020diffwave, chen2020wavegrad, mittal2021symbolic, huang2022prodiff, jeong2021diff, ulhaq2022efficient, avrahami2022blended, kim2022diffusionclip, bansal2023universal, saharia2022photorealistic, po2023state, zhang2023survey} and are one of the fundamental building blocks of image and audio synthesis systems, such as DALL-E \cite{rombach2022high}, stable diffusion \cite{ramesh2022hierarchical}, and Diffwave \cite{kong2020diffwave}.

Diffusion models' performance is appraised of high-fidelity sample generation and allows versatile guidance to control the generation. The simplest example of generation under guidance is to generate images of certain categories, such as cats or dogs. Such categorical information is taken as a conditional signal and fed into conditional diffusion models. \re{In more detail, we train conditional diffusion models using a labeled data set consisting of sample pairs $(x_i, y_i)$, where $y_i$ is the label of an image $x_i$. The training is to estimate a conditional score function using the data set, modeling the correspondence between $x$ and $y$. In this way, conditional diffusion models are learning the conditional distribution $P(x = \text{image} ~|~ y = \text{given~label})$ and allow sampling from the distribution}.

\begin{figure}[!htb]
\centering
\includegraphics[width = 0.72\textwidth]{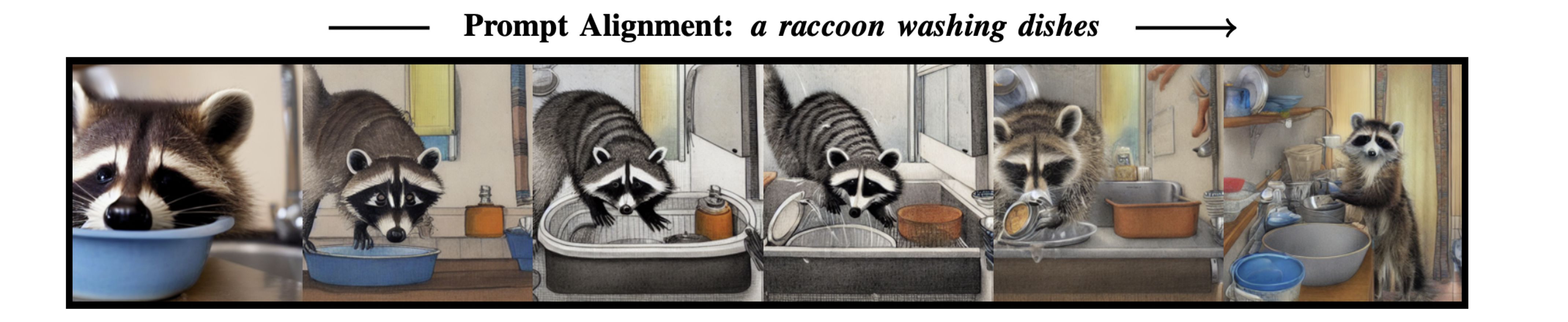}\\
\includegraphics[width = 0.7\textwidth]{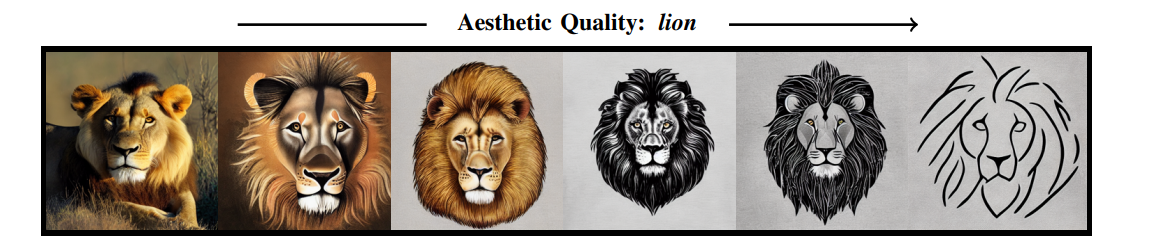}
\caption{Conditional diffusion models generate images under various guidance \cite{black2023training}. The upper row demonstrates an alignment with text description consisting of multiple objects. The lower row demonstrates an abstract description of aesthetic quality.}
\label{fig:text2image}
\end{figure}

In text-to-image synthesis systems, the conditional information is an input text prompt, which can be a sentence consisting of objects or more abstract requirements, e.g., aesthetic quality. \re{To generate images aligned with prompts, conditional diffusion models are trained with a massive annotated data set encompassing image and text summary pairs denoted as $(x_i, y_i)$. The text $y_i$ will be transformed into a word embedding and taken as input to a conditional diffusion model. Similar to the generation of images in certain categories, conditional diffusion models for text-to-image synthesis learn the conditional distribution $P(x = \text{image} ~|~y = \text{text~prompt})$ and allow sampling from it. In more sophisticated synthesis systems, some fine-tuning steps are implemented to further enable abstract prompt conditioning and improve the quality of generated images. For example, \cite{black2023training} reformulates the discretized backward process \eqref{eq:backward_sde} as a finite-horizon Markov Decision Process (MDP). The state space represents images, the conditional score function is viewed as a policy, and a reward function is defined to measure the alignment of an image to its desired text prompt. Therefore, to generate prompt-aligned images amounts to optimize reward via finding an optimal policy. \cite{black2023training} proposes a policy gradient-based method for fine-tuning pre-trained diffusion models. In Figure~\ref{fig:text2image}, we demonstrate a progressive improvement from left to right of fine-tuning a conditional diffusion model using the method in \cite{black2023training}}.

Conditional diffusion models are also a powerful tool in image editing and restoration \cite{wang2022zero, kawar2022denoising, choi2021ilvr, saharia2022image, saharia2022palette, lugmayr2022repaint, li2022srdiff, whang2022deblurring}, as well as audio enhancement \cite{lu2021study, welker2022speech, richter2023speech, yu2023conditioning}; see also surveys \cite{li2023diffusion, zhang2023survey} and the references therein. To showcase the idea, we consider the image inpainting task as an example. The goal of inpainting is to predict missing pixels of an image. We denote the known region of an image as $y$ and the original full image as $x$. Then inpainting boils down to sampling $x$ from the conditional distribution $P(x = \text{full~image} ~|~y = \text{known~region~of~the~image})$. In all these applications, conditional diffusion models are shown to be highly expressive and effective in modeling the conditional distributions \cite{song2020score, lugmayr2022repaint}.

\subsection{Control and Reinforcement Learning}\label{sec:control_application}

Apart from primary computer vision and audio tasks, diffusion models are actively deployed in Reinforcement Learning (RL) and control problems with appealing performance. For example, \cite{pearce2023imitating, chi2023diffusion, hansen2023idql, reuss2023goal, ding2023consistency} utilize conditional diffusion models to parameterize control/RL policies in highly complicated tasks, e.g., robot control and human behavior imitation. An extended review of the connection between diffusion models and RL can be found in \cite{zhu2023diffusion}. \re{In RL/control problems, a policy is a conditional probabilistic distribution on the action space given the state of an underlying dynamical system. Accordingly, when using diffusion models to parameterize policies, the goal is to learn a distribution $P(a = \text{action} ~|~ y = \text{system~states})$. \cite{pearce2023imitating, hansen2023idql} focus on the imitation learning scenario, where the goal is to mimic the behaviors of an expert. The data set contains expert demonstrations denoted by $(y_i, a_i)$ pairs. Here $y_i$ is the state of the system and $a_i$ is the expert's chosen action. Analogous to text-to-image synthesis, we train a conditional score network using the data set to capture the dependency between states and actions. During inference, given a new system state, we use the learned conditional diffusion model to generate plausible actions. Diffusion-QL \cite{wang2022diffusion} further adds regularization to the training of the conditional diffusion model and tries to learn optimal actions based on a pre-collected data set}.

\begin{figure}[!htb]
\centering
\includegraphics[width = 0.8\textwidth]{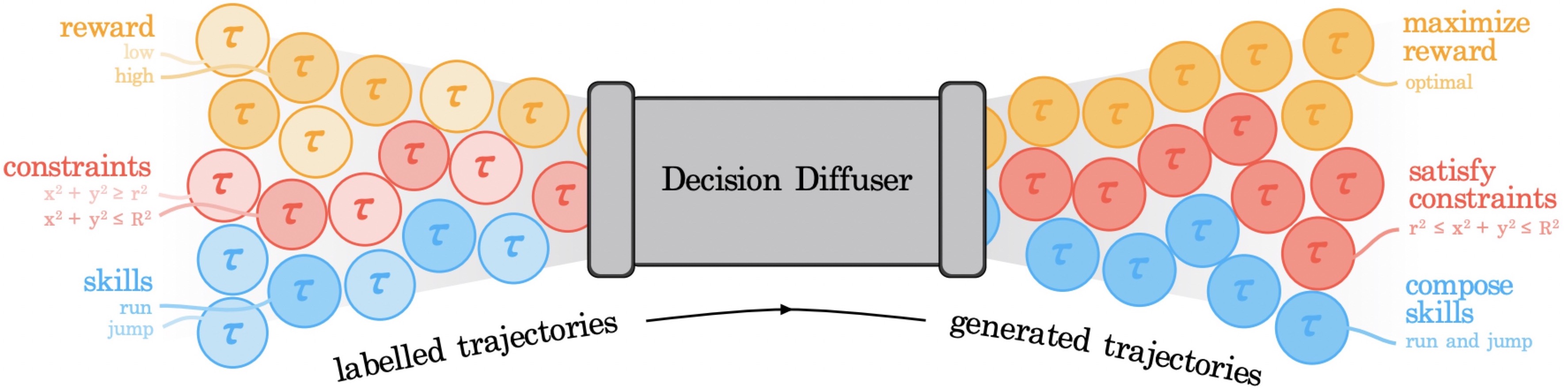}
\caption{Decision diffuser in \cite{ajay2022conditional}. The model is trained on labeled trajectories and is capable of generating state-action trajectories conditioned on desired reward values, constraints, or skills.}
\label{fig:decision_diffuser}
\end{figure}

Diffusion models also embody a new realm for algorithm design in control and RL problems by viewing sequential decision making as generative sequence modeling. \re{In a typical task of reward-maximization planning in RL, the goal is to find an optimal policy that achieves large accumulative rewards. Conventional methods rely on iteratively solving for the Bellman optimality to obtain a corresponding policy. Generative sequence modeling, however, directly produces state-action trajectories of large rewards, avoiding explicitly solving for Bellman optimality. In other words, generative sequence modeling directly samples from the conditional distribution $P(\tau = \text{state-action~trajectory} ~| ~ \tau~\text{attains~large~rewar})$}. Early success was demonstrated with transformer generative models \cite{janner2021sequence, chen2021decision}. Later, conditional diffusion models are deployed with state-of-the-art performance. Namely, Diffuser \cite{janner2022planning} generates state-action trajectories conditioned on high reward as guidance via conditional diffusion models. Decision Diffuser \cite{ajay2022conditional} presents conditional trajectory generation, taking reward, constraints, or skills as guidance and enhances Diffuser's performance. \re{For instance, given a pre-collected data set consisting of $(\tau_i, y_i)$, where $\tau_i$ is the state-action trajectory and $y_i$ is the accumulative reward of $\tau_i$. We use a conditional diffusion model to model the conditional distribution $P(\tau | y)$, by estimating the conditional score function. After training, we specify a proper target reward value and deploy the conditional diffusion model to generate sample trajectories. A policy can then be extracted from the generated trajectories via an inverse dynamics model \cite{agrawal2016learning}}. See the working flow of the decision diffuser in Figure~\ref{fig:decision_diffuser}. AdaptDiffuser \cite{liang2023adaptdiffuser} further introduces a discriminator for fine-tuning the conditional diffusion models, allowing self-evolution and adaptation to out-of-distribution tasks.

\subsection{Life-Science Applications}
In life-science applications, conditional diffusion models are making ever profound impacts \cite{anand2022protein, cao2022high, chung2022mr, chung2022score, gungor2023adaptive, jing2022torsional, lee2022proteinsgm, luo2022antigen, mei2022metal, waibel2022diffusion, ingraham2022illuminating, huang20223dlinker, schneuing2022structure, wu2022diffusion, gruver2023protein, weiss2023guided, xu2022geodiff, song2021solving, watson2023novo}. See also a survey \cite{guo2023diffusion} on applications of diffusion models in bioinformatics. These results cover diverse tasks including single-cell image analysis, protein design and generation, drug design, small molecule generation, etc. The performance surpasses many of their predecessors using autoregressive, VAE, or GAN-type deep generative models \cite{zhong2019reconstructing, zhong2021cryodrgn, shin2021protein, strokach2022deep}.

To demonstrate the use of conditional diffusion models, we take protein design as an example. Protein design can be posed as a problem of finding a sequence $w$ of certain length, where each coordinate of the sequence represents the structural information of the protein. A protein is only useful if it can be expressed in living cells. A widely adopted metric of usefulness is the likelihood of a protein sequence being a natural one \cite{gruver2023protein}. In addition, the binding affinity and aggregation
tendency are also vital properties of the protein structure. Combined with the usefulness metric, all these properties can be summarized by a vector-valued function $f(w)$. In this sense, conditional diffusion models actually generate protein sequences $w$ following a conditional distribution $P(w~|~ f(w) \in \cE)$, where $\cE$ is a set describing plausible protein structures. \re{The training of conditional diffusion models for protein generation is analogous to text-to-image diffusion models, based on a training data set containing diverse protein structures with measured properties. In the inference stage, we can first sample one configuration from $\cE$ and conditioned on the configuration, we generate new proteins.}

\subsection{Black-Box Optimization}\label{sec:application_opt}
In control, RL, and life-science applications, various guidance may be summarized as an abstract reward function $V(\cdot)$. Then the goal is to generate new samples from a conditional distribution, aiming to optimize the reward. Consequently, conditional diffusion models act as an optimizer which generates optimal solutions.

We revisit the example of offline reward-maximization planning in RL. Recall that our data set comprises state-action trajectories $\tau_i$ and the associated accumulative rewards $y_i = V(\tau_i) + \epsilon_i$, where $\epsilon_i$ is an independent observation noise. Reward-maximization planning essentially seeks solutions to the black-box optimization problem $\argmax_{\tau} V(\tau)$. \re{In this setting, we are prohibitive to interact with the target function $V$ beyond the given data set \cite{trabucco2022design}. Early existing works utilize GANs for optimal solution generation \cite{kumar2020model}, yet suffer from training instability and mode collapse issues. Recently, \cite{krishnamoorthy2023diffusion} empirically presents superior performance of generating high-quality solutions using conditional diffusion models. The idea is to transform the black-box optimization problem into a conditional sampling problem. In specific, given a proper target value $a$, conditional diffusion models generate solutions from the conditional distribution $P(\tau ~|~ V(\tau) = a)$. The subtlety stems from how to properly choose the target value $a$ to ensure the high-quality of the generated solutions. Roughly speaking, we are luring to choose a large $a$ so that the generated solutions achieve large rewards. However, if we choose $a$ too large compared to the given data set, significant extrapolation is required to generate corresponding solutions, leading to potential quality degradation. Consequently, a proper choice on $a$ heavily depends on the coverage of the collected data set. \cite{yuan2023reward} provides theoretical guidelines on how to choose $a$ to ensure good generated solutions, which we will introduce in Section~\ref{sec:cdm_opt}. Empirically, \cite{krishnamoorthy2023diffusion} proposes several methods to encourage large-reward solutions during the training of the conditional diffusion model, such as sample reweighting --- assigning large weights to samples with large rewards.}

\section{Theoretical Progress on Unconditional Diffusion Models}\label{sec:theory_uncondition}
This section reviews recent progress in the theoretical understanding of diffusion models. We recall from Section~\ref{sec:pre} that the score function is the key to implementing a diffusion model. From a theoretical perspective, the performance of diffusion models is intimately tied to whether or not the score function can be learned accurately. For a systematic treatment, we first introduce methods for learning the score and then dive into their theoretical insights. Specifically, we discuss how to properly choose neural networks for learning the score function, based on the universal and adaptive approximation capability of neural networks. More importantly, we demonstrate structural properties in the score function induced by data distribution assumptions, e.g., low-dimensional support and graphical models. Then we provide statistical sample complexities for estimating the score using the chosen neural networks. We are particularly interested in understanding how score estimation circumvents the curse of dimensionality issues in high-dimensional settings. Lastly, we study statistical rates for estimating the data distribution.

\subsection{Learning Score Function}\label{sec:score_estimation}

We consider the goal of learning the score function $\nabla \log p_t(x_t)$ using neural networks. A naive objective function is the weighted quadratic loss:
\begin{align}\label{eq:score_conceptual}
\min_{s \in \cS} \int_{0}^T w(t) \EE_{x_t \sim P_t} \left[\norm{\nabla \log p_t(x_t) - s(x_t, t)}_2^2 \right] \diff t,
\end{align}
where $w(t)$ is a weighting function and $\cS$ is a concept class (deep neural networks). However, such an objective function is not computable using samples, as the score function $\nabla \log p_t$ is unknown. As shown in the seminal works \cite{hyvarinen2005estimation} and \cite{vincent2011connection}, rather than minimizing the integral \eqref{eq:score_conceptual}, we can minimize an equivalent objective function,
\begin{align}\label{eq:denoising_score_matching}
\min_{s \in \cS} & \int_{0}^T w(t) \EE_{x_0 \sim P_{\rm data}} \Big[\EE_{x_t \sim {\sf N}(\alpha(t)x_0, h(t)I_D)}\Big[\big\|\nabla_{x_t} \log \phi_t(x_t | x_0) - s(x_t , t)\big\|_2^2 \Big] \Big]\diff t.
\end{align}
Here, $\phi_t(x_t | x_0)$ denotes the Gaussian transition kernel of the forward process, so that $\nabla \log \phi_t$ admits an analytical form
\[
    \nabla_{x_t} \log \phi_t(x_t | x_0) = -\frac{x_t - \alpha(t)x_0 }{h(t)}.
\]
By this analytical expression, we could approximate the objective \eqref{eq:denoising_score_matching} using finite samples. Note that $\nabla_{x_t} \log \phi_t(x_t | x_0)$ is the noise added to $x_0$ at time $t$. Therefore, \eqref{eq:denoising_score_matching} is also known as the denoising score matching. As shown in \cite{huang2021variational, luo2022understanding}, denoising score matching can also be derived using a variational perspective, reproducing the evidence lower bound for regularized data negative likelihood minimization.

\paragraph{Score Blowup and Early-Stopping} One challenge of optimizing \eqref{eq:denoising_score_matching} is the score blowup issue \cite{vahdat2021score, song2020improved}. To demonstrate the phenomenon, we consider a data distribution which lies in a linear subspace, where $x = Az$ for a representation matrix $A \in \RR^{D \times d}$ and a latent variable $z \in \RR^d$. \re{Here $D$ represents the ambient dimension of data and $d$ is the intrinsic dimension, which is often much smaller than $D$.} As shown in \cite{chen2023score}, the ground truth score $\nabla \log p_t(x)$ assumes the following orthogonal decomposition,
\begin{align}\label{eq:score_decomp}
\nabla \log p_t(x) = A \nabla \log p_t^{\rm ld}(A^\top x) + \underbrace{\frac{1}{1-e^{-t}} (I - AA^\top) x}_{(\cT)},
\end{align}
where $p_t^{\rm ld}$ is the marginal density function of applying the forward diffusion process \eqref{eq:forward_sde} on the latent variable $z$. As can be seen, the term $(\cT)$ is orthogonal to the subspace spanned by matrix $A$. More importantly, as $t$ approaches $0$, the magnitude of $(\cT)$ grows to infinity as long as $x \neq 0$. The reason behind this is that $(\cT)$ enforces the orthogonal component to vanish so that the low-dimensional subspace structure is reproduced in generated samples. Such a blowup issue appears in all geometric data \cite{vahdat2021score}. As a consequence, an early stopping time $t_0 > 0$ is introduced and the practical score estimation loss is written as
\begin{align}\label{eq:score_practical}
\min_{s \in \cS} & \int_{t_0}^T w(t) \EE_{x_0 \sim P_{\rm data}} \Big[\EE_{x_t \sim {\sf N}(\alpha(t)x_0, h(t)I_D)}\Big[\big\|\nabla_{x_t} \log \phi_t(x_t | x_0) - s(x_t , t)\big\|_2^2 \Big] \Big]\diff t.
\end{align}
For practical implementation, we approximate \eqref{eq:score_practical} by its empirical version. Specifically, given $n$ i.i.d. data points $x_i \sim P_{\rm data}$ for $i = 1, \dots, n$, we sample $x_t$ given $x_0 = x_i$ from the Gaussian distribution ${\sf N}(\alpha(t) x_i, h(t)I_D)$. We also sample time $t$ from the interval $[t_0, T]$ to approximate the integration with respect to $t$.

\begin{figure}[!htb]
\centering
\includegraphics[width = 0.7\textwidth]{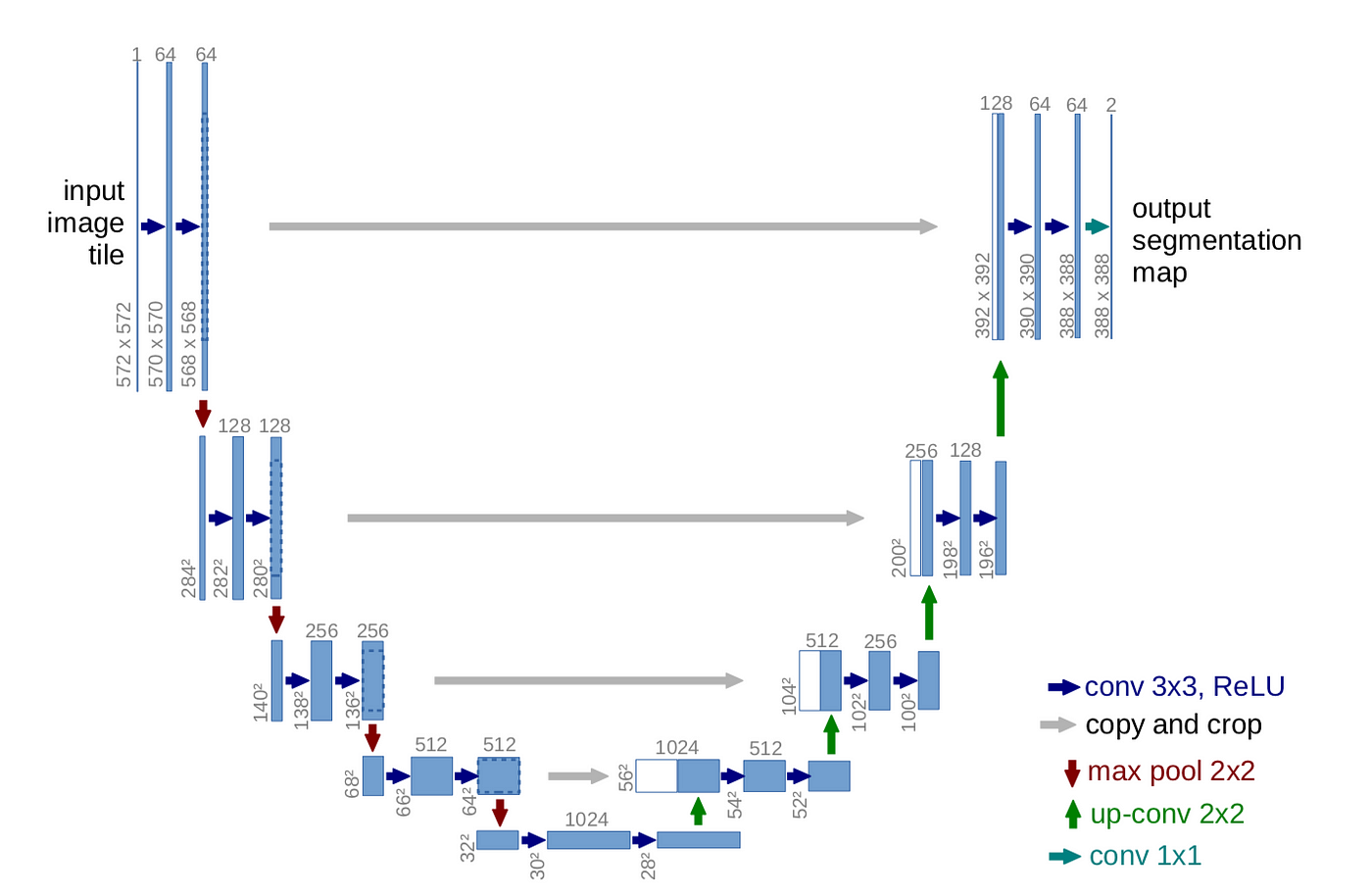}
\caption{U-Net architecture in \cite{ronneberger2015u} for $32 \times 32$ resolution RGB images. \re{When generating new samples using a discretized backward process, diffusion models utilize the U-Net at each discretization step for transforming samples. The image sample together with a time embedding is first compressed into a low-dimensional representation and then lifted back to the original dimension.}}
\label{fig:unet}
\end{figure}

\begin{remark}[Network class $\cS$]
A common choice of the network class $\cS$ is the U-Net \cite{ronneberger2015u}, as demonstrated in Figure~\ref{fig:unet}. The network architecture utilizes convolution layers and shortcut connections. In the network, an input is first compressed into a low-dimensional representation and then gradually lifted back to the original dimension. This encoder-decoder type structure aims to extract intrinsic structures in data and leads to an efficient learning. More recently, using transformer-based score network has demonstrated outstanding performance \cite{peebles2023scalable,gupta2023photorealistic,liu2024sora}, which excels in capturing spatial-temporal dependencies in data.
\end{remark}

\subsection{Score Approximation and Estimation}\label{sec:score_guarantee}
The choice of concept class $\cS$ is vital to the learning of the score function as in \eqref{eq:score_practical}. There are two requirements on $\cS$: 1) Class $\cS$ should be rich enough to well-approximate the ground truth score function, i.e., there exists a candidate in $\cS$ close to $\nabla \log p_t$; 2) Class $\cS$ should not be overly complicated to obscure the learning process with finite training samples. We present theoretical insights on both aspects, and address 1) from a function approximation perspective and 2) from a statistical learning perspective. 

\subsubsection{Score Approximation Guarantees}
The question underscores score approximation is what the size and the architecture of score networks to ensure the existence of an $\epsilon$-error approximation to the score function. Here $\epsilon > 0$ is the desired error level and often represents an $L^2$ distance measure. Such a question is reminiscent of the universal function approximation ability of neural networks \cite{cybenko1989approximation, hornik1990universal, barron1994approximation, guhring2020error, yarotsky2018optimal, lu2021deep, schmidt2020nonparametric, chen2019efficient}. However, we highlight some fundamental differences between score approximation and conventional function approximation. Firstly, the score function is defined on all of the high-dimensional Euclidean space, due to the added Gaussian noise, while conventional neural network approximation theory focuses on compact domains. Secondly, the score function depends on an additional time dimension, which complicates its approximation.

Concurrent works \cite{oko2023diffusion} and \cite{chen2023score} tackle the challenges via very different approaches and develop score approximation theories for Euclidean data and low-dimensional linear subspace data. \cite{oko2023diffusion} rewrites the score function as $\nabla \log p_t = \frac{\nabla p_t}{p_t}$ and uses neural networks to approximate $p_t$ and $\nabla p_t$ separately. To address the time dependency, \cite{oko2023diffusion} proposes a series of ``diffused basis functions''. More formally, diffused basis functions are the convolutions of the Gaussian transition kernel in the forward process \eqref{eq:forward_sde} with time-independent polynomials, such as Taylor polynomials and B-splines. The idea behind the diffused basis functions can be understood as tracking the evolution of $p_t$ with respect to time $t$. Indeed, once we can approximate the density of the clean data distribution $P_{\rm data}$ with time-independent polynomials, the corresponding diffused polynomials automatically approximate the density $p_t$ for all $t$.

\begin{figure}[!htb]
\centering
\includegraphics[width = 0.65\textwidth]{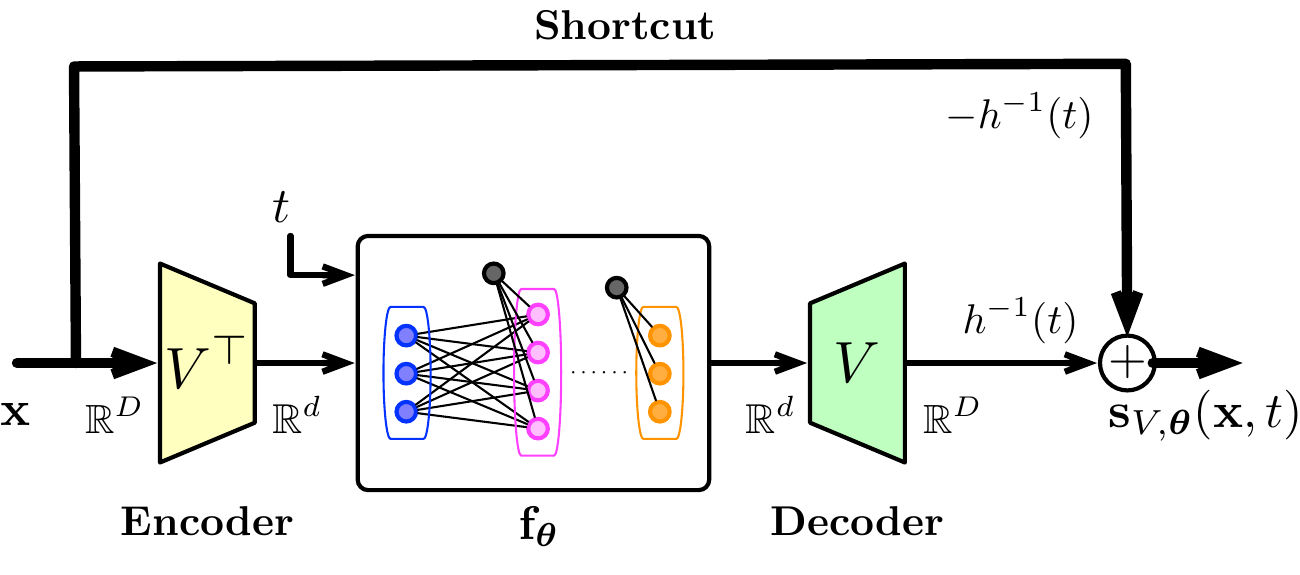}
\caption{Simplified U-Net architecture in \cite{chen2023score} for approximating score functions in the low-dimensional subspace data setting. Matrix $V$ represents the linear encoder and decoder, which is to be jointly learned with parameter $\theta$ during the optimization of loss \eqref{eq:score_practical}. Here $f_\theta$ is a network with input and output dimensions being the subspace dimension.}
\label{fig:simplified_scorenet}
\end{figure}

On the other hand, \cite{chen2023score} resorts to a local Taylor approximation of the score function using neural networks. In this case, the score function $\nabla \log p_t$ is viewed as a multi-dimensional input-output mapping of certain regularity. Built upon the existing universal approximation theories of neural networks, \cite{chen2023score} devises a score approximation result. More interestingly, \cite{chen2023score} considers low-dimensional linear subspace data and shows that the ground truth score $\nabla \log p_t$ decomposes into two terms as in \eqref{eq:score_decomp}. In this regard, a simplified U-Net architecture (Figure~\ref{fig:simplified_scorenet}) with linear encoder and decoder is constructed for efficient score approximation, indicating that the data subspace structures circumvent the dependence on data ambient dimension.

In deriving the approximation guarantees, \cite{oko2023diffusion, chen2023score} leverage sophisticated input truncation to deal with the unbounded domain. The approximation error is in turn measured in the $L^2$ norm sense, instead of commonly used $L^\infty$ norm. In order to achieve an $\epsilon$ approximation error, the network size scales in the order $\tilde{\cO}(\epsilon^{-\gamma})$, where $\gamma$ is data dimension dependent. We emphasize that when there exists low-dimensional subspace structures in data, $\gamma$ only depends on the subspace dimension.

\subsubsection{Sample Complexity of Score Estimation}
We shift to understand how many samples are needed to learn a score estimator by optimizing \eqref{eq:score_practical}. The learned estimator should generalize in the sense that its deviation to the ground truth score is small. This requires more than a good score network class $\cS$, but also the learnability within $\cS$, which is characterized by some complexity measure of $\cS$.

An early work \cite{block2020generative} provides a sample complexity bound for score estimation. Yet the bound depends on some unknown Rademacher complexity of the score network class. Built upon the score approximation theory, \cite{oko2023diffusion} and \cite{chen2023score} establish score estimation theories from the nonparametric statistics point of view. \cite{oko2023diffusion} assumes the clean data distribution is supported on a unit cube with a Besov continuous density. In order to obtain an $\epsilon$-accurate score estimator in the $L^2$ norm, the sample size grows in the order $\tilde{\cO}(\epsilon^{-\frac{D + 2\beta}{\beta}})$, where $D$ is the data dimension and $\beta$ is the smoothness index of the density. As can be seen, the sample complexity indicates the curse of dimensionality, and \cite{oko2023diffusion} reduces the dependence on $D$ when data has a known linear subspace structure. In the independent study, \cite{chen2023score} focuses on linear subspace data without knowing the subspace in advance. Under assumptions of data having Gaussian tail and the score being Lipschitz, \cite{chen2023score} establishes a $\tilde{\cO}(\epsilon^{-(d + 5)})$ sample complexity, where $d$ is the subspace dimension. While free of the curse of dimensionality, \cite{chen2023score} also proves that the unknown subspace can be automatically estimated via score estimation. Turning towards a kernel-based approach, \cite{yang2022convergence, wibisono2024optimal} establish optimal statistical score estimation rates using a regularized Gaussian kernel density estimation method. The obtained sample complexity is $\tilde{\cO}(\epsilon^{-(d+4)})$ for Lipschitz continuous score functions.

\paragraph{Optimization Guarantees on Score Estimation} On the algorithmic side, we are aware of \cite{shah2023learning} studying score estimation in Gaussian mixture models. They provide convergence analysis of using gradient descent to minimize the score estimation loss \eqref{eq:score_practical}. The algorithmic behavior can be characterized in two phases, where in the large-noise phase, i.e., time $t$ large in \eqref{eq:score_practical}, gradient descent is analogous to power iteration. In the the small-noise phase, i.e., $t$ small, gradient descent is akin to the EM algorithm. Besides, \cite{han2024neural} studies the optimization guarantee of using two-layer neural networks for score estimation.

\subsubsection{Score Estimation in Graphical Models}\label{sec:graphical_model}
Besides considering data distributions in continuous spaces, such as Euclidean space and linear subspace, \cite{mei2023deep} studies score approximation and estimation in graphical models. Graphical models such as Markov random fields and restricted Boltzmann machines have been widely used for modeling image distributions in the literature \cite{geman1986markov, ranzato2010factored}, yet they are fundamentally different from distributions on continuous variables. \cite{mei2023deep} proposes a novel approach for controlling the sample complexity of score estimation in high dimensions. In particular, the authors view the neural networks in diffusion models as a denoising algorithm, enabling an efficient score approximation.

Specifically, \cite{mei2023deep} assumes the data distribution follows an Ising model $P_{\rm data}(\sigma) \propto \exp\{ \langle \sigma, A \sigma\rangle \}$ for $\sigma \in \{ \pm 1\}^D$. Under certain high-temperature conditions of the matrix $A \in {\mathbb R}^{D \times D}$, the score function $s(x_t, t)$ can be approximately computed by variational inference algorithms such as message passing \cite{mezard2009information, eldan2018gaussian, celentano2023local}. This is an iterative algorithm of the form $m^{\ell+1} = \tanh( (A - K) m^{\ell} + c_t x_t)$ with $m^0 = 0$ for certain matrix $K \in {\mathbb R}^{D \times D}$. Each step of the message-passing algorithm comprises simple operations, including matrix-vector multiplication and pointwise nonlinearity, which could be efficiently approximated by one block of the residual network $u^{\ell + 1} = u^{\ell} + W_1{\rm ReLU}(W_2 u^{\ell})$. This renders an efficient approximation of Ising model score functions using a residual network with $\cO(D^2 L)$ parameters, where $L$ is the number of neural network layers, allowing a moderate dependence on the problem size. Incorporating standard Rademacher complexity generalization error bound, \cite{mei2023deep} provides an estimation error bound without the exponential dependence on dimensionality.

\subsection{Sampling and Distribution Estimation}\label{sec:distro_guarantee}

Our ultimate goal of diffusion models is to learn the data distribution and provide easy access to generating new samples. This section first reviews sampling theories of diffusion models via the backward process \eqref{eq:backward_practice}, with a basic assumption on the accuracy of the estimated score function $\hat{s}$. Next, we move to an end-to-end analysis of diffusion models, by presenting sample complexity bounds for learning distributions. 

\subsubsection{Sampling Theory} Several recent sampling theories of diffusion models prove that the distribution generated by the backward process is close to the data distribution, as long as the score function is accurately estimated. The central contribution is a relationship between $\epsilon_{\rm dis}$ and $\epsilon_{\rm score}$, where $\epsilon_{\rm dis}$ is a discrepency between the sampled data distribution and the ground truth distribution, and $\epsilon_{\rm score}$ is the score estimation error. Specifically, \cite{de2021diffusion, albergo2023stochastic} establish upper bounds of $\epsilon_{\rm dis}$ using $\epsilon_{\rm score}$ for diffusion Schr\"{o}dinger bridges. The error $\epsilon_{\rm dis}$ is measured in the total variation distance and $\epsilon_{\rm score}$ is measured in the $L^\infty$ norm. More concrete bounds of $\epsilon_{\rm dis}$ are provided in \cite{block2020generative, lee2022convergencea, chen2022sampling, lee2022convergenceb, benton2024nearly}. These works specialize $\epsilon_{\rm score}$ to be the $L^2$ error of the estimated score function, and $\epsilon_{\rm dis}$ to be the total variation distance between the generated distribution and the data distribution. \cite{lee2022convergencea} requires the data distribution satisfying a log-Sobolev inequality. Concurrent works \cite{chen2022sampling} and \cite{lee2022convergenceb} relax the log-Sobolev assumption on the data distribution to only having bounded moments. The upper bound in \cite{chen2022sampling} takes the form
\begin{align*}
\epsilon_{\rm dis} = \tilde{\cO}\left(\sqrt{T}\epsilon_{\rm score} + \textsf{discretization-error} + \textsf{forward-error} \right).
\end{align*}
Here $T$ is the terminal time in the forward process. The {\sf discretization-error} depends on the regularity of the data distribution and the step size in the discretized backward process. The {\sf forward-error} quantifies the divergence between $P_T$ and $P_\infty = {\sf N}(0, I_D)$, since the forward process is terminated as a finite time $T$. It is worth mentioning that \cite{lee2022convergenceb} allows $\epsilon_{\rm score}$ to be time-dependent and \cite{benton2024nearly} improves the data dimension dependency. Recently, \cite{chen2023restoration, chen2023probability, benton2023linear, li2024accelerating, li2023towards} largely enrich the study of sampling theory using diffusion models. Specifically, novel analyses based on Taylor expansions of the discretized backward process \cite{li2023towards} or localization method \cite{benton2023linear, benton2024nearly} are developed, which improve the upper bound on $\epsilon_{\rm dis}$. Further, \cite{chen2023restoration} extends to DDIM sampling scheme and \cite{chen2023probability} considers the probabilistic ODE backward sampling.

Besides Euclidean data, \cite{de2022convergence} makes the first attempt to analyze diffusion models for learning low-dimensional manifold data. Assuming $\epsilon_{\rm score}$ is small under the $L^\infty$ norm (extension to the $L^2$ norm is also provided), \cite{de2022convergence} bound $\epsilon_{\rm dis}$ of diffusion models in terms of the Wasserstein distance. The obtained bound has an exponential dependence on the diameter of the data manifold. Moreover, \cite{montanari2023posterior} considers using diffusion processes to sample from noisy observations of symmetric spiked models and \cite{el2023sampling} studies polynomial-time algorithms for sampling from Gibbs distributions based on diffusion processes. The construction of diffusion processes in \cite{montanari2023posterior, el2023sampling} leverages the idea of stochastic localization \cite{eldan2013thin, montanari2023sampling, chen2022localization, el2022information}.

\paragraph{Computational efficiency of sampling through diffusion models} Sampling from certain high-dimensional distributions can be computationally challenging. For instance, \cite{el2022sampling} demonstrates the hardness of sampling from the low-temperature Sherrington-Kirkpatrick model using any stable algorithms. An intriguing line of inquiry would be to understand the computational complexity of sampling through diffusion models and its connection to the complexity of sampling via the Langevin dynamics. 

Using heuristic physics methods, \cite{ghio2023sampling} investigated the relationship between the computational complexity of sampling through the Langevin dynamics and diffusion models in high-dimensional distributions widely studied in the statistical physics of disordered systems. They utilized the hardness of computing the score function as a proxy for the hardness of sampling with diffusion models. They generated phase diagrams of the computational complexity of sampling from these high-dimensional models, and identified parameter regions where diffusion models are unable to sample efficiently, while the Langevin dynamics can; conversely, they also identified regions where the Langevin dynamics are inefficient, yet diffusion models perform well.

\subsubsection{Sample Complexity of Distribution Estimation}  Distribution estimation theory of diffusion models is explored in \cite{song2020sliced} and \cite{liu2022let} from an asymptotic statistics point of view. These results do not provide an explicit sample complexity bound. Given the aforementioned sampling theory and score estimation theory, we can develop an end-to-end analysis of diffusion models. The following theorem summarizes the existing sample complexity bounds of diffusion models in \cite{oko2023diffusion} and \cite{chen2023score}.
\begin{theorem}[Sample Complexity of Distribution Estimation]\label{thm:distribution_estimation}
Suppose the data distribution $P_{\rm data}$ is supported on a cube $[-1, 1]^D$ with a density function of smoothness index $s$. Under some conditions\footnote{These are technical conditions on the data distribution approaching the boundary of the hypercube. A precise statement can be found in \cite{oko2023diffusion}.}, diffusion models can learn a distribution $\hat{P}$ satisfying
\begin{align}\label{eq:highD_distribution_estimation}
d_{\rm TV}(\hat{P}, P_{\rm data}) = \tilde{\cO}\left(n^{-\frac{s}{2s + D}} \right),
\end{align}
where $d_{\rm TV}$ is the total variation distance. 

Moreover, suppose the data distribution is supported on a $d$-dimensional subspace, i.e., data $x = Az$ with an unknown matrix $A \in \RR^{D \times d}$ of orthonormal columns. \re{We recall that $d$ is the intrinsic dimension and much smaller than $D$. Specializing the smoothness index $s = 1$ and under some conditions, diffusion models can estimate the subspace and learn a distribution $\hat{P}_{\rm sub}$ in the subspace satisfying}\footnote{The precise statement requires a set of notations. Interested readers may refer to \cite[Theorem 3]{chen2023score}.}
\begin{align}\label{eq:lowD_distribution_estimation}
d_{\rm TV}(\tilde{P}_{\rm data}, \hat{P}_{\rm sub}) = \tilde{\cO}\left(n^{-\frac{1}{d + 5}}\right),
\end{align}
where $\tilde{P}_{\rm data}$ is a slightly perturbed data distribution.
\end{theorem}
From \eqref{eq:highD_distribution_estimation}, we conclude that if the density function has a higher smoothness $s$, the distribution estimation is more efficient. Moreover, \eqref{eq:highD_distribution_estimation} matches the minimax optimal rate of distribution estimation in Euclidean spaces, indicating diffusion models are powerful and efficient distribution estimators. The result in \eqref{eq:lowD_distribution_estimation} further unveils the adaptivity of diffusion models, since the convergence rate is only dependent on the subspace dimension $d$, which can be much smaller than $D$. This result provides valuable insights of why diffusion models yield startling practical performance, since real-world high-dimensional data often has rich low-dimensional geometric structures and diffusion models are efficient in capturing these structures for an efficient learning.

\subsection{Alternative Formulation: Stochastic Localization}

Stochastic localization is a measure-valued stochastic process employed to study isoperimetric inequalities \cite{eldan2013thin, eldan2020taming, eldan2022analysis}. As a mathematical technique, stochastic localization has been successfully utilized in proving versions of the Kannan-Lovász-Simonovits (KLS) conjecture \cite{lee2016eldan, chen2021almost}. The process was later generalized in \cite{el2022sampling, montanari2023posterior, alaoui2023sampling} as a sampling algorithm with provable sampling error bounds. The connections between stochastic localization and the DDPM (Denoising Diffusion Probabilistic Model) diffusion models are demonstrated in \cite{montanari2023sampling}.

We introduce the simplest stochastic localization process, following the presentation in \cite{montanari2023sampling}. Given the measure $P_{\rm data}$, the stochastic localization process is a stochastic differential equation defined as:
\begin{align}\label{eqn:stochastic-localization}
\diff Y_t & = m_t(Y_t) \diff t + \diff W_t \quad \text{for} \quad t \in [0, \infty),~~~ Y_0 = 0,
\end{align}
where $m_t(y) = \EE_{(x, g) \sim P_{\rm data} \otimes {\sf N}(0, I_D)}[x ~|~ t x + \sqrt{t}g = y ]$ is the posterior expectation of $Y_t$ upon observing $y = t x + \sqrt{t}g$. Standard theory implies that the marginal distribution of $Y_t$ satisfies $Y_t \stackrel{d}{=} t x + \sqrt{t} g$, where $(x, g) \sim P_{\rm data} \otimes {\sf N}(0, I_D)$. Consequently, $\lim_{t \to \infty} Y_t / t$ converges to a random variable following the distribution $P_{\rm data}$. In generative modeling tasks, one could fit the posterior expectation $m_t(y)$ using neural networks and training samples, and discretize the SDE as in Eq.~(\ref{eqn:stochastic-localization}), similar to DDPM diffusion models.

In sampling tasks for the distribution $P_{\rm data}$ being spin-glasses models and posterior of spiked matrix models, \cite{el2022sampling, montanari2023posterior, alaoui2023sampling} show that the posterior expectation $m_t$ can be approximately computed using variational inference algorithms in the high-temperature regime, enabling efficient sampling from these distributions. 

A firm connection between stochastic localization to DDPM diffusion models is shown in \cite{montanari2023sampling}: The stochastic localization process $\{ Y_t \}_{t \ge 0}$ as in Eq. (\ref{eqn:stochastic-localization}) is equivalent to the backward SDE of the diffusion model \eqref{eq:backward_sde} up to time and scale reparametrization. \cite{montanari2023sampling} further generalizes the stochastic localization scheme to general stochastic processes. 

\section{Theoretical Progress on Conditional Diffusion Models}\label{sec:theory_cdm}
Although conditional diffusion models share many characteristics with their unconditional counterpart, their unique reliance on guidance requires new understanding and insights. As a result, theoretical results on conditional diffusion models are highly limited. In this section, we mimic the study of unconditional diffusion models yet put an extra emphasis on distinct uses and methods of conditional diffusion models. We first introduce the training of conditional diffusion models, which is to estimate the conditional score function. Interestingly, the conditional score function can be related to the unconditional score function, motivating a fine-tuning perspective for training conditional diffusion models. Next, we present conditional score estimation and distribution estimation guarantees. The last section is devoted to theoretical insights on the influence of guidance in Gaussian mixture models, where we corroborate common observations and reveal curious new discoveries.

\subsection{Learning Conditional Score}\label{sec:conditional_score}
For conditional sample generation via \eqref{eq:conditional_backward}, the conditional score function $\nabla \log p_t(x | y)$ needs to be estimated. We slightly abuse the notation to denote $s$ as a conditional score network and $\cS$ as the corresponding network class. By introducing an early-stopping time $t_0$, a conceptual quadratic loss for conditional score estimation is defined as
\begin{equation}\label{eq:conditional_score_matching}
\argmin_{s \in \cS} \int_{t_0}^T w(t) \EE_{(x_t, y)} \left[\norm{\nabla \log p_t(x_t | y) - s(x_t, y, t)}_2^2\right] \diff t,
\end{equation}
where $w(t)$ is a time dependent reweighting function. Inspired by \cite{hyvarinen2005estimation} and \cite{vincent2011connection}, \cite[Proposition 3.1]{li2024diffusion} asserts the equivalence of \eqref{eq:conditional_score_matching} to the following implementable loss function,
\begin{align}\label{eq:conditional_score_practical}
\argmin_{s \in \cS} \int_{t_0}^T \EE_{(x_0, y)} \left[ \EE_{x_t \sim {\sf N}(\alpha(t)x_0, h(t)I_D)}\left[\norm{\nabla_{x_t} \log \phi_t(x_t | x_0) - s(x_t, y, t)}_2^2\right] \right]\diff t,
\end{align}
which shares the same spirit as \eqref{eq:backward_practice}.

\paragraph{Classifier and Classifier-Free Guidance}
Practical implementations of conditional score estimation, such as classifier and classifier-free guidance methods, build upon \eqref{eq:conditional_score_practical} for reduced computational cost or better performance \cite{dhariwal2021diffusion,ho2022classifier}. We begin with the classifier guidance method \cite{dhariwal2021diffusion}, which is arguably the first method to allow conditional generation in diffusion models similar to GANs or flow models \cite{brock2018large, kingma2018glow}. Specifically, when conditional information $y$ is discrete, e.g., image categories, the conditional score $\nabla \log p_t(x_t | y)$ is rewritten via Bayes' rule as
\begin{align*}
\nabla \log p_t(x_t | y) = \nabla \log p_t(x_t) + \nabla \log c_t(y | x_t),
\end{align*}
where $c_t$ is the likelihood function of an external classifier. In other words, classifier guidance combines the unconditional score function with the gradient of an external classifier. The external classifier is trained using the diffused data points in the forward process. As a result, the performance of classifier guidance methods is sometimes limited, since it is difficult to train the external classifier with highly corrupted data.

Later, classifier-free guidance proposes to remove the external classifier, circumventing the limitation caused by classifier training. The idea of classifier-free guidance is to introduce a mask signal to randomly ignore $y$ and unifies the learning of conditional and unconditional scores. Specifically, let $\tau \in \{\emptyset, {\rm id}\}$ be a mask signal, where $\emptyset$ means to ignore the conditional information $y$ and ${\rm id}$ to keep $y$. Corresponding to the two circumstances, we have
\begin{align}
\tau = \emptyset: & \quad \int_{t_0}^T \EE_{(x_0, y)} \left[\EE_{x_t \sim {\sf N}(\alpha(t)x_0, h(t)I_D)} \left[\norm{s(x_t, \emptyset, t) - \nabla_{x_t} \log \phi_t(x_t | x_0)}_2^2\right]\right] \diff t \nonumber \\
\tau = {\rm id}: & \quad \int_{t_0}^T \EE_{(x_0, y)} \left[\EE_{x_t \sim {\sf N}(\alpha(t)x_0, h(t)I_D)} \left[\norm{s(x_t, y, t) - \nabla_{x_t} \log \phi_t(x_t | x_0)}_2^2\right]\right] \diff t. \label{eq:label_score}
\end{align}
Note that \eqref{eq:label_score} coincides with \eqref{eq:conditional_score_practical}, and recall that $t_0$ is an early-stopping time. Combining the two cases, classifier-free guidance method minimizes the following loss function:
\begin{align}\label{eq:classifierfree_guidance}
\hat{s} \in \argmin_{s \in \cS} \int_{t_0}^T \EE_{(x_0, y)} \left[\EE_{\tau \sim P_\tau, x_t \sim {\sf N}(\alpha(t)x_0, h(t)I_D)} \left[\norm{s(x_t, \tau y, t) - \nabla_{x_t} \log \phi_t(x_t | x_0)}_2^2\right]\right] \diff t.
\end{align}
\re{Here $\tau$ is randomly chosen among $\emptyset$ and ${\rm id}$ following distribution $P_\tau$. The simplistic choice on $P_\tau$ is a uniform distribution on $\{\emptyset, {\rm id}\}$, while it is preferred to bias towards $\tau = {\rm id}$ in some applications \cite{ho2022classifier}}.

Once the estimator $\hat{s}$ is learned from \eqref{eq:classifierfree_guidance}, we compute
\begin{align}\label{eq:guidance_strength}
\tilde{s}(x, y, t) = (1+\eta) \cdot \hat{s}(x, y, t) - \eta \cdot \hat{s}(x, \emptyset, t)
\end{align}
with some $\eta > 0$ for substitution into the backward process. From a theoretical point of view, choosing $\eta > 0$ is counter-intuitive, as the resulting $\tilde{s}$ does not correspond to the conditional score function $\nabla \log p_t(x | y)$. However, \re{a properly chosen $\eta$ leads to improved performance on benchmarks in practice. More interestingly, increasing $\eta$ reduces the diversity of the generated samples but promotes distinguishablity of them \cite{ho2022classifier}.} The coefficient $\eta$ can also be chosen dependent on time $t$. Unfortunately, a principled guidance on how to choose $\eta$ is still missing, yet some theoretical insights on the impact of $\eta$ are developed \cite{wu2024theoretical}; see also Section~\ref{sec:strength_impact}.

\paragraph{Adapting Uncondtional Score via Guidance}
In real use cases, the desired criteria or objectives of conditional sample generation may shift over time, which necessitates quick adaptation of conditional diffusion models. Although classifier-free guidance method has been adopted for training conditional diffusion models from scratch, it is not tailored for adapting or fine-tuning diffusion models owing to the computational overhead. Consequently, this opens up new possibilities of theories and methods for fine-tuning diffusion models without compromising the pre-training performance.

Recently, \cite{clark2023directly, black2023training, fan2023dpok, uehara2024feedback, tang2024fine} propose efficient fine-tuning methods when the quality of generated samples is measured by a scalar-valued reward function. To guide a pre-trained model for generating high-reward samples, \cite{clark2023directly} assumes the differentiability of the reward function and directly fine-tunes parameters in the diffusion model by back-propagation. \cite{black2023training, fan2023dpok} formulate the sample generation process of diffusion models as a finite-horizon Markov decision process. The score function is equivalent to a policy and allows for fine-tuning using reinforcement learning techniques, such as policy gradient methods.

A more interesting and principled fine-tuning method draws motivation from the classifier guidance. We revisit the Bayes' rule for conditional score function,
\begin{align*}
\nabla \log p_t(x_t | y) = \underbrace{\nabla \log p_t(x_t)}_{\textrm{pre-trained score}} + \underbrace{\nabla \log c_t(y | x_t)}_{\rm guidance},
\end{align*}
where the classifier $c_t$ acts as guidance to adapting the pre-trained score. Despite classifier guidance requires a discrete label $y$ (yet can be multi-dimensional), the decomposition in the last display has a profound impact on guidance-based fine-tuning. Indeed, \cite{janner2022planning, bansal2023universal, uehara2024fine} extend guidance to arbitrary conditioning by incorporating gradients of a proper scalar-valued function. For demonstration, \cite{bansal2023universal} defines the so-call ``universal guidance'' in the form of $\nabla_{x_t} \ell(y, f(\hat{x}_0))$, where $f$ is a function measuring the quality of samples, $\hat{x}_0$ is the anticipated generated sample of the pre-trained diffusion model given current point $x_t$ in the backward process, and $\ell$ is a loss function. Note that $\hat{x}_0$ correlates with $x_t$ and the gradient is nontrivial. As a special example, when $y$ is the discrete label, $f$ is the classification likelihood, and $\ell$ is the cross-entropy loss, universal guidance reproduces classifier guidance. 

\subsection{Conditional Score and Distribution Estimation}\label{sec:cdm_guarantee}
Theory of conditional score estimation and conditional distribution estimation is very limited. To the best of our knowledge, \cite{li2024diffusion} provides an initial study of using \eqref{eq:conditional_score_practical} for conditional score estimation and distribution estimation. A systematic analysis of classifier-free guidance method is presented in \cite{fu2024unveil}, with results highlighted by approximation theories of conditional score functions and sample complexities of conditional score estimation and distribution learning. In addition, \cite{fu2024unveil} shows the utility of the developed statistical theory in elucidating the performance of conditional diffusion models for diverse applications, including model-based transition kernel estimation in reinforcement learning, solving inverse problems \cite{chung2022diffusion, rout2024solving, chung2022improving,kawar2022denoising}, and reward conditioned sample generation.

The core contribution of \cite{fu2024unveil} is the conditional score approximation theory, which is motivated by the idea of diffused basis approximation in \cite{oko2023diffusion}. In more detail, \cite{fu2024unveil} substantially broadens the framework to unbounded data domains and conditional distributions. The authors rewrite the conditional score function as $\nabla \log p_t(x | y) = \frac{\nabla p_t(x| y)}{p_t(x | y)}$ and approximate $\nabla p_t(x| y)$ and $p_t(x|y)$ separately. On a technical side, the unbounded data domain and the conditioning on $y$ lead to new challenges. More importantly, however, \cite{fu2024unveil} lifts the technical conditions on data distributions in \cite{oko2023diffusion} and obtains optimal statistical rates with a mild bounded H\"{o}lder norm assumption. We remark that \cite{fu2024unveil} takes the condition $y$ as independent input variables, leaving an open direction to identify intrinsic smoothness with respect to $y$ in the conditional distribution so as to improve the dimension dependency.

\subsection{Theoretical Insights on Strength of Guidance}\label{sec:strength_impact}
We conclude the discussion on conditional diffusion models by a recent work on the influence of the strength of guidance \cite{wu2024theoretical}. We are referring back to \eqref{eq:guidance_strength} and studying the influence of $\eta$ on the sample generation. The same strength parameter can be introduced into classifier guidance as
\begin{align*}
\tilde{s}(x, y, t) = \nabla \log p_t(x_t) + \eta \nabla \log c_t(y | x_t).
\end{align*}
Hence, we will not distinguish different guidance methods, and term $\eta$ as the strength of guidance.

\begin{figure}[!htb]
    \centering
    \includegraphics[width = 0.9\textwidth]{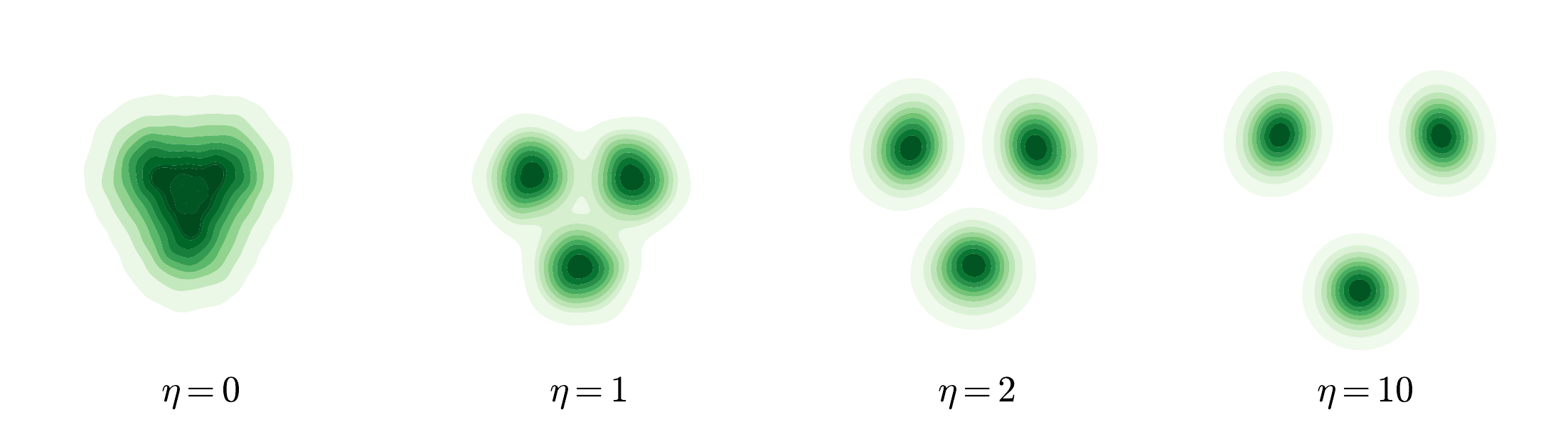}
    \caption{The effect of guidance strength $\eta$ on a three-component GMM in $\RR^2$ \cite{ho2022classifier, wu2024theoretical}. Each component has weight $1 / 3$ and identity covariance, and the component centers are $(\sqrt{3} / 2, 1 / 2)$, $(-\sqrt{3} / 2, 1 / 2)$ and $(0, -1)$. The leftmost panel displays the unguided density. We increase the guidance strength from left to right. \re{When generating samples, we use the ground truth score function.}}
    \label{fig:strength_3GMM}
\end{figure}

A common observation of the consequence yielded by $\eta$ is best illustrated in Figure~\ref{fig:strength_3GMM} on a three-component Gaussian mixture model. Here label $y$ indicates the Gaussian components and $x$ is a two-dimensional variable. When generating new samples, we fix a choice on $y$ to obtain within component samples. We observe that with an increased guidance strength $\eta$, the generated conditional distribution shifts its probability mass farther away from other components, and most of the mass becomes concentrated in smaller regions. 

The results in \cite{wu2024theoretical} theoretically characterize the influence of strength on diffusion models in the context of Gaussian mixture models. Under mild
conditions, \cite{wu2024theoretical} proves that incorporating strong guidance not only boosts classification confidence but also
diminishes distribution diversity, leading to a reduction in the differential entropy of the generated conditional distribution. These theories align closely with empirical observations.

\begin{figure}[!htb]
    \centering
    \includegraphics[width = 0.8\textwidth]{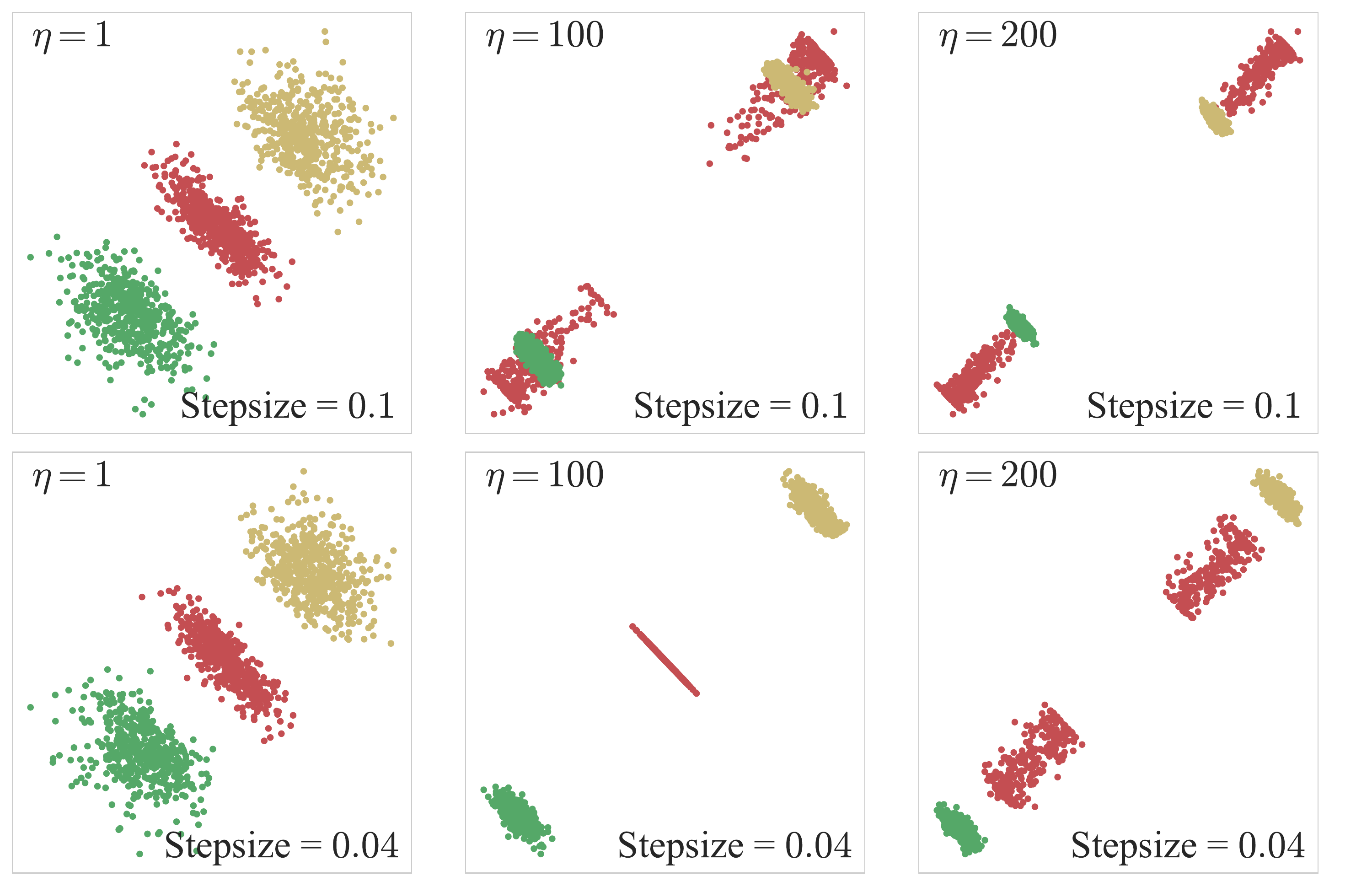}
    \caption{Illustration of a possible negative effect of large guidance strength in \cite{wu2024theoretical}. In this plot, the component means of the Gaussian mixture model are aligned on the same line. We increase the guidance strength $\eta$ from left to right. The upper row uses a relatively large discretization step size in the backward process. With large $\eta$, the center component splits into two clusters at an earlier stage. The bottom row uses a much smaller discretization step size; the center component then splits only with a much larger $\eta$.}
    \label{fig:negative}
\end{figure}

On the other hand, \cite{wu2024theoretical} identifies a possible negative impact of large $\eta$ under discretized backward
sampling in Gaussian mixture models, as depicted in Figure~\ref{fig:negative}. There exists a phase shift as strength $\eta$ increases. Under large $\eta$, the center component of the original Gaussian mixture model splits into two symmetric clusters, harming the modality of the original data. The emergence of this negative effect is tied to the locations of the components and the discretization step size in the backward sampling process. Until this point, we are not aware of principled methods for tuning the strength $\eta$ in different tasks, which might be encouraged by the obtained theoretical insights.

\section{Diffusion Model for Optimization}\label{sec:cdm_opt}
This section introduces a novel avenue for optimization in high-dimensional complex and structured spaces through diffusion models. We focus on data-driven black-box optimization, where the goal is to generate new solutions that optimize an unknown objective function. Black-box optimization, also known as model-based optimization in machine learning, encapsulates various application domains such as reinforcement learning, computational biology and business management \cite{watson2023novo,guo2023diffusion,ajay2022conditional,pan2020novel,pan2019user,luenberger1984linear, fu2021offline,bu2023offline}.

Solving data-driven black-box optimization differentiates from conventional optimization, as interactions with the objective function beyond a pre-collected data set are prohibitive, diminishing the possibility of sequentially searching for optimal solutions. Instead, people aim to extract pertinent information from the pre-collected data set and directly recommend solutions. To complicate matters, the solution space is often high-dimensional with rich latent structures. For example, in drug discovery, molecule structures need to satisfy global and local regularity to be expressive in living bodies. This poses a critical requirement for solving data-driven black-box optimization: we need to capture data latent structures to avoid suggesting unrealistic solutions that deviate severely from the original data domain.

\begin{figure}[!htb]
    \centering
    \includegraphics[width = 0.95\textwidth]{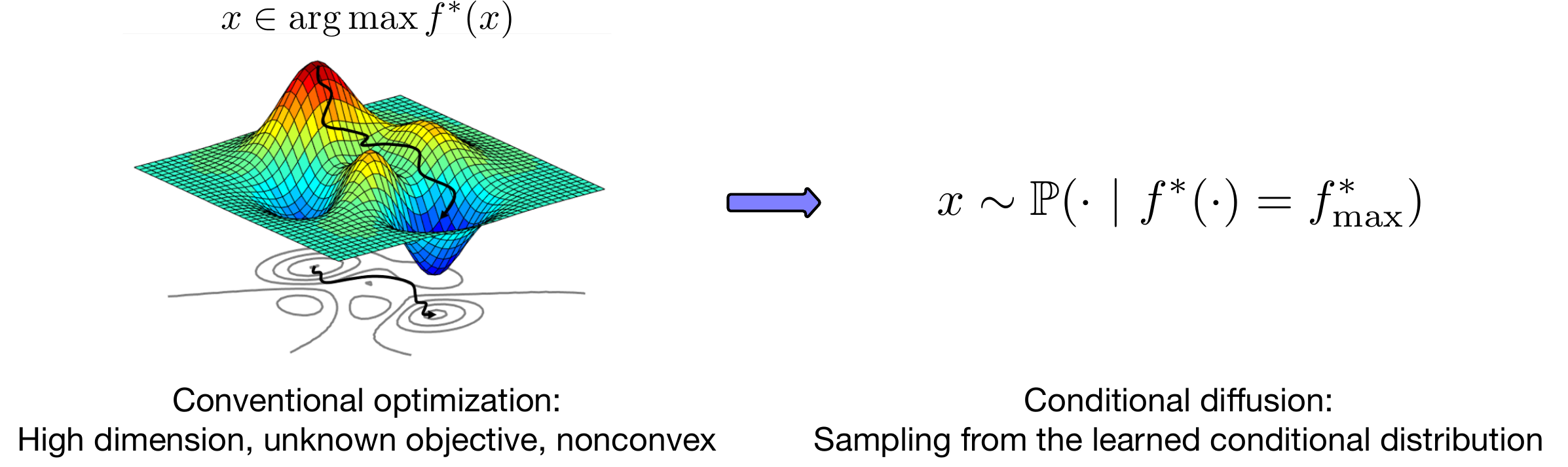}
    \caption{Reformulation of data-driven black-box optimization as conditional sampling in \cite{li2024diffusion}. The conditional distribution is learned from a pre-collected data set.}
    \label{fig:generative_opt}
\end{figure}

To address the challenges, \cite{li2024diffusion} formulates data-driven black-box optimization as sampling from a conditional distribution, as demonstrated in Figure~\ref{fig:generative_opt}. The objective function value is the conditioning in the conditional distribution, meanwhile the distribution implicitly captures data latent structures.

The pre-collected data set in \cite{li2024diffusion} consists of two parts: 1) a massive unlabeled part $\cD_{\rm unlabel}$ and 2) a smaller labeled part $\cD_{\rm label}$. By terming the objective function as a reward function, \cite{li2024diffusion} considers two types of label feedback in $\cD_{\rm label}$:
\begin{itemize}
\item {\bf (Real-valued reward)} The data set $\cD_{\rm label}$ consists of data and reward pairs, where the reward is a real-valued noise-perturbed version of the underlying ground truth reward;
\item {\bf (Human preference)} The data set $\cD_{\rm label}$ consists of triples taking two comparable data points and a binary preference label. The preference label indicates that the corresponding data point is likely to have an edge in the underlying reward over the other one.
\end{itemize}
Moreover, the data point $x \in \RR^D$ is assumed to concentrate on a linear subspace, i.e., $x = Az$ for some unknown matrix $A \in \RR^{D \times d}$, with $z \in \RR^d$ being the latent variable. Therefore, newly generated samples should be kept close to the subspace to maintain high fidelity. 

\begin{figure}[!htb]
\centering
\includegraphics[width = 0.95\textwidth]{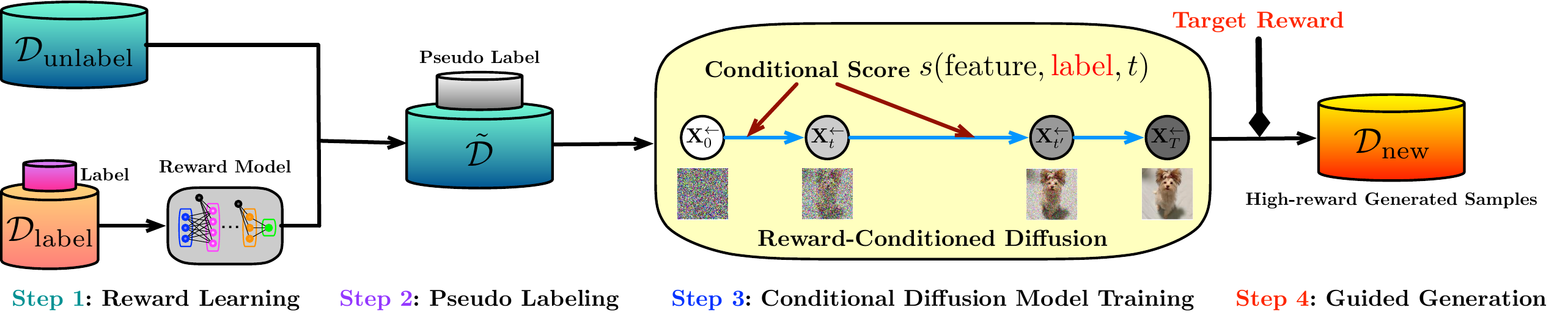}
\caption{The learning algorithm proposed in \cite{li2024diffusion} consists of four steps. In the first step, a reward model is learned from the labeled data $\cD_{\rm label}$. In the second step, the learned reward model is deployed as a pseudo-labeler to label $\cD_{\rm unlabel}$. In the third step, a conditional diffusion model is trained using the pseudo-labeled data. Lastly, in the fourth step, new samples are generated from the conditional distribution $P_a$ by specifying a target reward value $a$.}
\label{fig:cdm4opt}
\end{figure}

A semi-supervised learning algorithm is proposed in Figure~\ref{fig:cdm4opt}. There are two training procedures: one in the first step for estimating the reward function and the other one in the third step for training the conditional diffusion model. In the fourth step, the target reward is set at a scalar value $a$, so that the generated samples follow the conditional distribution $\hat{P}_a = \hat{P}(\cdot ~|~ \hat{\textrm{reward}} = a)$, where $\hat{P}$ and $\hat{\textrm{reward}}$ emphasize that the distribution and the reward are estimated, rather than the ground truth. One may be curious about the quality of the generated samples. In particular, these two properties of the generated samples are of particular interest: 1) the reward levels of new samples and 2) their level of fidelity -- how much do new samples deviate from the latent subspace.

The results in \cite{li2024diffusion} provide a positive statistical answer. For the reward levels of new samples, \cite{li2024diffusion} defines
\begin{align*}
\texttt{SubOpt}(a) = a - \EE_{x \sim \hat{P}_a} [V(x)]
\end{align*}
to measure the gap between the sample average reward and the target reward. In the language of
bandit learning, $\texttt{SubOpt}$ can be interpreted as a form of off-policy sub-optimality. The following theorem proves an upper bound on $\texttt{SubOpt}$.
\begin{theorem}[Conditional Diffusion for Black-Box Optimization]
\label{thm:cdm4opt}
Consider data $x = Az$ for some unknown matrix $A \in \RR^{D \times d}$ with orthonormal columns. Suppose the reward function $V$ decomposes into
\begin{align*}
V(x) = \underbrace{g(AA^\top x)}_{\geq 0, \textrm{~on-support reward}} + \underbrace{h((I - AA^\top)x)}_{\leq 0, \textrm{~off-support penalty}}.
\end{align*}
Running the algorithm in Figure~\ref{fig:cdm4opt} generates high-fidelity samples and gives rise to
\begin{align*}
\texttt{SubOpt}(a) \leq \underbrace{\EE_{P_a}\left[| g - \hat{g} |\right]}_{\textrm{reward~estimation~error}} + \underbrace{\left| \EE_{P_a}[g] - \EE_{\hat{P}_a} [g] \right|}_{\textrm{on-support~diffusion~error}} + \underbrace{\left|\EE_{\hat{P}_a}[h]\right|}_{\textrm{off-support~penalty}},
\end{align*}
where $\hat{g}$ is an estimated reward function and $P_a = P(\cdot ~|~ \hat{\textrm{reward}} = a)$.
\end{theorem}
We note that the reward function $V$ consists of two components: 1) The on-support reward $g$ is nonnegative and measures the quality of samples by projecting it onto the subspace spanned by matrix $A$; 2) The off-support penalty, however, is nonpositive and discourages the generated samples extrapolating in the space outside the subspace spanned by matrix $A$.

Theorem~\ref{thm:cdm4opt} says that the reward estimation error depends on the sample size in $\cD_{\rm label}$, which is often the dominating term. The on-support diffusion error and off-support penalty depend on the sample size in $\cD_{\rm unlabel}$ and rely on a statistical analysis of conditional diffusion models for distribution estimation. There is also a subtlety to explicitly quantify the three error terms, namely, the distribution shift, which is the mismatch between the training data distribution and the target data distribution. Diffusion models are learned to generate similar samples to the training distribution, however, optimizing the reward function drives the model to deviate from the training. In other words, the model needs to both ``interpolate" and ``extrapolate". A higher value of $a$ provides stronger guidance to the diffusion model, while the increasing distribution shift may hurt the generated samples' quality.

Through detailed analysis, \cite{li2024diffusion} instantiates Theorem~\ref{thm:cdm4opt} to parametric and nonparametric settings. For example, with a linear reward function $g$, the reward estimation error aligns with the optimal off-policy bandit sub-optimality \cite{nguyen2021offline, jin2021pessimism}, where the distribution shift is explicitly computed and the dimension dependence is $d$ instead of large ambient dimension $D$. In the human preference setting, \cite{li2024diffusion} considers the Bradley-Terry-Luce choice model \cite{bradley1952rank,zhu2023principled} and derives a similar concrete sub-optimality bound.

\section{Future Directions}\label{sec:open_question}
We discuss several future directions of diffusion models, exploring their connections to stochastic control and distributional robustness; we also introduce discrete diffusion models.

\subsection{Connection to Stochastic Control}
In either unconditioned diffusion models or conditional diffusion models, generating samples using backward processes \eqref{eq:backward_sde} or \eqref{eq:conditional_backward} can be viewed as a stochastic control problem \cite{aastrom2012introduction, fleming2012deterministic}. The goal of stochastic control is to design the evolution of the controlled variable, so that certain cost is minimized. In diffusion models, the score function constitutes the control and steers the quality of the generated samples. In the simplest form of unconditioned diffusion models, we define the cost to be the distribution divergence between the generated distribution and the data distribution, such as the total variation distance and the Wasserstein distance. Then the score estimation essentially amounts to finding the optimal control for minimizing such costs.

When using conditional diffusion models for black-box optimization, the cost is the negative of a reward function and the conditional score function is the control. The theory in \cite{li2024diffusion} chooses a proper target reward to design the control for optimizing the cost. Leveraging this control perspective, a series of empirical results attempt to fine-tune diffusion models by designing the control based on various cost forms \cite{clark2023directly, lee2023aligning, wu2023better, black2023training, fan2023dpok, xu2023imagereward, hao2022optimizing, watson2021learning, wallace2023end, lou2023reflected}. For instance, \cite{clark2023directly, fan2023dpok, black2023training} consider differentiable real-valued reward, while \cite{wu2023better, xu2023imagereward} focus on the cost being human preferences. In terms of methodology, \cite{black2023training} uses the policy gradient method in reinforcement learning for fine-tuning the control (conditional score function). \cite{clark2023directly} resorts to the classifier guidance formula by directly augmenting the unconditioned score function by gradients of the cost.

In this regard, principled methodologies and accompanying theories \cite{uehara2024feedback, uehara2024fine, tang2024fine} can be motivated from the stochastic control perspective, improving and analyzing diffusion models under various task objectives.

\subsection{Adversarial Robustness and Distributionally Robust
 Optimization}
Diffusion models exhibit the natural denoising property in the backward processes, which are leveraged for adversarial purification and promoting robustness \cite{nie2022diffusion, wu2022guided, carlini2022certified, xiao2022densepure}. To illustrate, in robust classification, a two-step classification procedure is proposed: A trained conditional diffusion model is first deployed to generate new samples given the input adversarial examples for multiple times, hoping to purify the added noise in the input sample. Then the generated samples are fed into a trained classifier to produce a predicted label. Due to the randomness in the diffusion models, multiple transformed samples of the same input adversarial example can be obtained. Therefore, a majority vote among the predicted labels is assigned as the label of the adversarial example. This method is motivated by a justification on the promotion of robustness using diffusion models and empirically shown to be effective \cite{xiao2022densepure}. Yet an end-to-end analysis is still missing.

We also expect a close connection between diffusion models to Distributionally Robust Optimization (DRO) \cite{delage2010distributionally, kuhn2019wasserstein, goh2010distributionally, rahimian2019distributionally}. As shown in Theorem~\ref{thm:distribution_estimation}, diffusion models generate samples in the close vicinity of a target distribution, which can be viewed as providing a certain coverage of the distributional uncertainty set in DRO. In this sense, diffusion models can potentially simulate the worst-case scenario in the uncertainty set. We suspect the emergence of innovative methods and theories in the corresponding intersection area, where motivating attempts have been made in \cite{xu2024flow}.

\subsection{Discrete Diffusion Models}
Discrete diffusion models, analogous to the previous continuous counterparts, are designed to keep the finite data support during the forward and backward processes \cite{meng2022concrete, campbell2022continuous, benton2022denoising, santos2023blackout, lou2023discrete, austin2021structured, sun2022score, hoogeboom2021argmax}. Instead of using continuous Gaussian noise to corrupt clean data, discrete diffusion resorts to continuous-time Markov processes for transforming clean data. The discrete nature has appealing alignment to real data characterized by a massive but finite support, e.g., natural language represented by word tokens and molecular structures. As reported in \cite{lou2023discrete}, discrete diffusion achieves competitive or better performance in language tasks with comparable sized models. \cite{li2024distribution} demonstrates the possibility of using discrete diffusion for solving combinatorial problems.

We describe a discrete distribution by a probability vector $p_{\rm data}$ belonging to the simplex. Analogous to  Gaussian noise corruption for continuous diffusion, we utilize a continuous-time Markov process driven by a time-dependent transition matrix $Q_t$, i.e.,
\begin{align}\label{eq:discrete_forward}
\frac{\diff p_t}{\diff t} = Q_t p_t \quad \text{with}\quad  p_0 = p_{\rm data}.
\end{align}
The process above is known as the forward discrete diffusion process. Several design choices of $Q_t$ are summarized in \cite{austin2021structured}, including discretized Gaussian, uniform, and absorbing transitions.

The discrete forward process \eqref{eq:discrete_forward} also asserts a time reversal:
\begin{align}\label{eq:discrete_backward}
\frac{\diff p_t^{\leftarrow}}{\diff t} = \bar{Q}_t p_t^{\leftarrow} \quad \text{with} \quad [\bar{Q}_t]_{ij} = \begin{cases}
\frac{[p_{_{T-t}}]_i}{[p_{_{T-t}}]_j} [Q_{T-t}]_{ji} & \text{if}~i \neq j \\
- \sum_{s \neq i} [Q_{T-t}]_{is} \frac{[p_{_{T-t}}]_s}{[p_{_{T-t}}]_i} & \text{if}~i = j
\end{cases}.
\end{align}
Here $\bar{Q}_t$ is the backward transition matrix and $[\cdot]_i$ (or $[\cdot]_{ij}$) denotes the $i$-th (or $(i, j)$-th) entry. We observe from the backward process \eqref{eq:discrete_backward} that to generate new samples, we only need to estimate the ratios $\frac{[p_t]_i}{[p_t]_j}$ for $t \in [0, T]$. We can view this probability ratio as an analogy to the score function in the continuous distribution. Some caveats arise that estimating the ratios suffers from the massive support size of the data distribution and the magnitude of ratios can vary significantly. It is also likely that a large fraction of the ratios are zero or approximately zero, inducing sparse structures. There are different empirical methods for estimating the ratios \cite{meng2022concrete, campbell2022continuous, lou2023discrete, austin2021structured, sun2022score}, such as using a quadratic loss or an entropy loss.

From a theoretical stand of point, discrete diffusion poses interesting open questions: How to efficiently estimate the ratios using finite samples, with potential sparse structures and ill-spread ranges of ratios. More importantly, how to smartly design principled transition kernels relevant to data distributions remains unclear. Nonetheless, assuming access to estimated ratios, \cite{chen2024convergence} proves the first sampling theory of discrete diffusion models.

\section{Conclusion}\label{sec:conclusion}
In this paper, we have surveyed how diffusion models generate samples, their wide applications, and existing theoretical underpinnings of them. We have adopted a continuous-time description of the forward and backward processes in diffusion models and discussed their training procedure, especially when there exists guidance to steer the sample generation. We have started with an exposure to theories of unconditional diffusion models, covering its score approximation, statistical estimation, and sampling theories. Built upon the insights from the unconditional diffusion models, we have then turned towards conditional diffusion models, with a focus on their unique design properties and theories. Next, we have made a connection between generative diffusion models to black-box optimization, paving a new avenue for high-dimensional optimization problems. Lastly, we have discussed several trending future directions.

\newpage
\bibliographystyle{IEEEtran}
\bibliography{ref}

\end{document}